%% file: main.tex
\documentclass[runningheads]{llncs}

\usepackage{eccv}

\usepackage{eccvabbrv}

\usepackage{graphicx}
\usepackage{booktabs}

\usepackage[accsupp]{axessibility}

\usepackage{array}
\usepackage{makecell}
\usepackage{multirow}
\usepackage{wrapfig}

\usepackage[colorlinks,citecolor=eccvblue]{hyperref}

\usepackage{orcidlink}
\usepackage{bbding}

\usepackage{placeins}
\usepackage{listings}
\usepackage{xcolor}
\begin{document}

\title{Anchoring on Reality: Breaking the Pseudo-Target Ceiling in Makeup Transfer}

\titlerunning{Reality-Anchored Makeup Transfer}

\newcommand{\projectlead}{\textsuperscript{\ensuremath{\dagger}}}
\newcommand{\corrauth}{\,{\Envelope}\,}

\author{
Bo Wei\inst{1}\orcidlink{0009-0000-4094-2156},
Xianhui Lin\inst{2}\projectlead\orcidlink{0000-0002-8974-2064},
Yi Dong\inst{2},
Zhongzhong Li\inst{2}\orcidlink{0000-0003-0570-3197},
Zonghui Li\inst{2} \\
Zirui Wang\inst{2},
Jiachen Yang\inst{2},
Xing Liu\inst{2},
Hong Gu\inst{2}\orcidlink{0009-0006-9939-0388} \\
Xiaoming Li\inst{3}\corrauth\orcidlink{0000-0003-3844-9308},
and Wangmeng Zuo\inst{1}\corrauth\orcidlink{0000-0002-3330-783X}
}

\authorrunning{B.~Wei et al.}

\institute{
\begin{tabular}{c}
$^{1}$Harbin Institute of Technology \quad
$^{3}$Nanjing University \\
$^{2}$vivo BlueImage Lab, vivo Mobile Communication Co., Ltd \\
\email{csbowei@gmail.com, xhlin129@gmail.com, ydong@outlook.com,} \\
\email{csxmli@nju.edu.cn, wmzuo@hit.edu.cn} \\
{\href{https://csbowei.github.io/ART/}{\textcolor[RGB]{255, 26,145}{https://csbowei.github.io/ART/}}}
\end{tabular}
}

\maketitle

\begingroup
\renewcommand{\thefootnote}{}
\footnotetext{\projectlead Project lead.
\corrauth Corresponding author.}
\endgroup

\begin{center}
    % \setlength{\abovecaptionskip}{4pt}
    % \setlength{\belowcaptionskip}{-6pt}
    % \vspace{-6pt}
    \includegraphics[width=\linewidth]{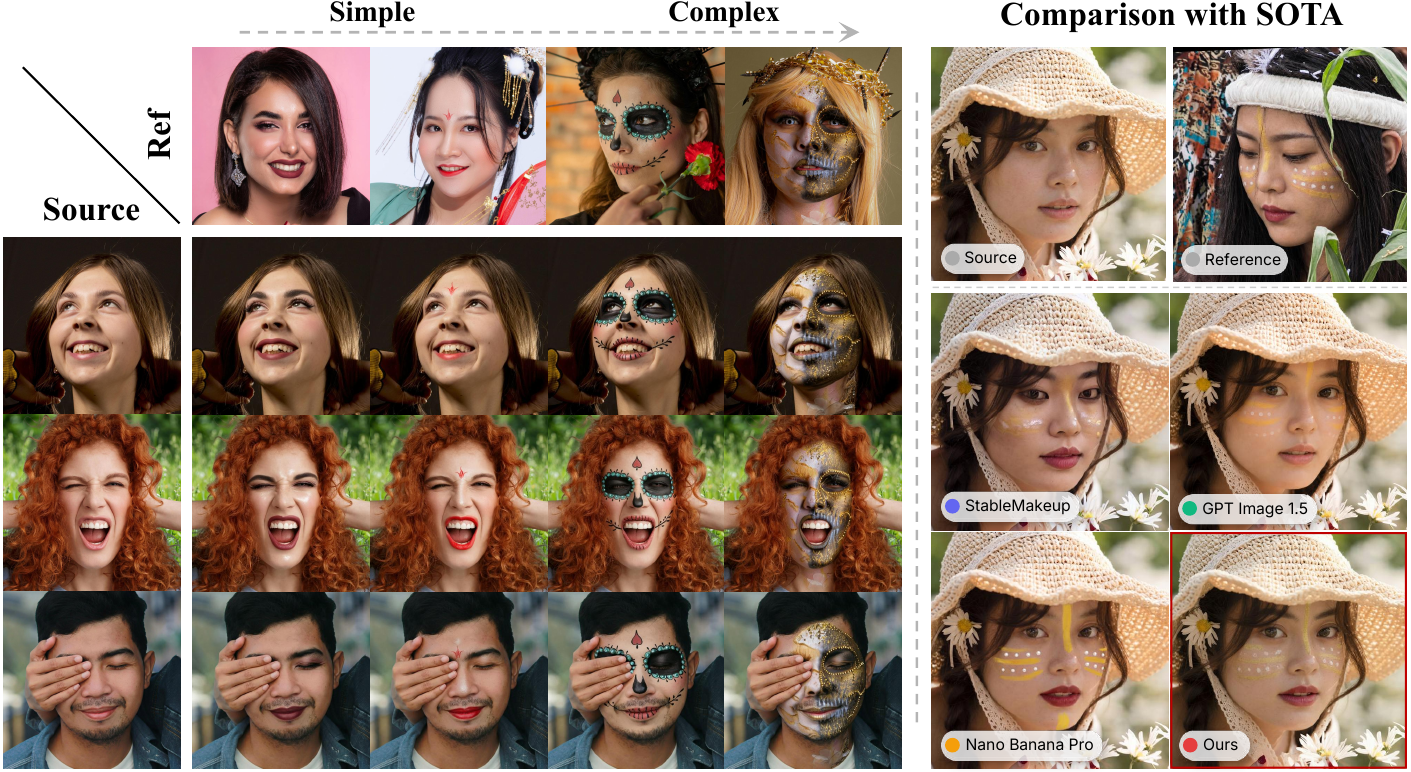}
    \captionof{figure}{ART~enables high-fidelity transfer of diverse makeup styles (simple to complex) at up to 2K resolution, while preserving identity, geometry, and makeup placement under challenging expressions, occlusions, and cross-gender scenarios.}
    \label{fig:teaser}
\end{center}

\begin{abstract}
Makeup transfer applies a reference cosmetic style to a source face while preserving its identity and geometry.
However, this task is severely hindered by the lack of real paired training data.
Current methods rely on either weak priors or synthetic pseudo-targets from large-scale editing models.
These paradigms provide suboptimal guidance, often leading to degraded fine-grained details, synthetic artifacts, and identity drift.
To this end, we propose \textbf{A}nchoring on \textbf{R}eality Makeup \textbf{T}ransfer~(ART), a two-stage framework with a reality-anchored refinement cycle.
In Stage I, the model is initialized with pseudo-targets to establish basic semantic alignment and global makeup placement.
Crucially, Stage II shifts supervision from pseudo-targets to the real reference, reconstructing it from its bare-skin counterpart through a differentiable cycle that penalizes any omitted detail and overrides synthetic artifacts.
Furthermore, we introduce MakeupFaces2K (MF2K), the first 2K-resolution in-the-wild makeup portrait dataset comprising 8,573 images.
Extensive experiments demonstrate that our method achieves superior makeup fidelity, strong background stability, and robust identity preservation, especially for complex makeup styles.

% \vspace{-6pt}
\end{abstract}

\section{Introduction}
\label{sec:intro}
The task of makeup transfer aims to synthesize a target portrait that faithfully adopts the cosmetic style of a reference image while strictly preserving the identity and facial geometry of a source image.
Beyond consumer photo editing, the demand for high-fidelity transfer is increasing in professional applications, such as cinematic post-production and virtual avatars.
In these settings, cosmetics often go beyond daily makeup to encompass complex styles, including heavy artistic patterns, face paintings, glitter, and dense stickers.
Faithfully transferring these styles, especially across gender or ethnicity, remains a formidable challenge for existing methods.

Fundamentally, progress in complex makeup transfer is bottlenecked by the lack of real paired supervision.
Many prior methods formulate makeup transfer as an unpaired image-to-image translation problem guided by weak priors, such as cycle consistency or domain adversarial losses~\cite{zhu2017unpaired, chen2019beautyglow, Gu_2019_ICCV, sun2024content, sun2024shmt, ruan2025mad}.
However, these approaches often struggle to achieve photorealistic fidelity and precise spatial placement when dense correspondence is ambiguous.
Driven by the recent surge in large-scale generative models, state-of-the-art methods typically construct synthetic pseudo-targets using external editing engines~\cite{zhang2025stablemakeup, wu2025evomakeup} and train models to regress their predictions to these generated pairs.
While this paradigm improves overall realism, it imposes an inherent \textbf{pseudo-target ceiling}: the model systematically inherits the biases and artifacts of the constructed targets, such as identity drift, degraded fine-grained details, and unintended background edits.
Such flaws are inevitable, since a model trained merely to imitate pseudo-targets can never surpass them.

To break this ceiling, we propose ART, a two-stage Diffusion Transformer (DiT)~\cite{peebles2023scalable} framework that shifts the learning paradigm from synthetic regression to differentiable reality-anchored reconstruction.
Our key insight is that, although the ideal transfer target is unavailable, the real reference itself provides all the authentic makeup details that supervision requires.
Unlike traditional cycle consistency (e.g., CycleGAN~\cite{zhu2017unpaired}), which only enforces invertibility back to the source, our cycle reconstructs the real reference itself, providing direct supervision for the transferred makeup.

Specifically, Stage~I performs pseudo-target initialization to establish basic semantic alignment and global makeup placement.
In Stage~II, we introduce a reality-anchored refinement cycle.
We first predict a preliminary transfer output and retain it as a differentiable \textit{makeup carrier} in the latent space.
The DiT is then constrained to reconstruct the real reference image from its bare-skin counterpart (extracted via an auxiliary makeup remover), conditioned solely on this carrier.
Crucially, because this carrier remains a differentiable latent rather than a decoded image, the reconstruction gradient flows back into the initial transfer prediction.
This directly penalizes any omission of fine cosmetic detail, forcing the transfer prediction to reproduce the reference's true textures and overwrite the artifacts inherited from the pseudo-targets.
To stabilize this refinement cycle, we introduce a controlled-noise bottleneck during the generation of the \textit{makeup carrier}.
By modulating the amount of pseudo-target prior retained under noise injection, this bottleneck preserves the coarse makeup placement as a structural foundation while affording the model sufficient flexibility to recover high-fidelity details from the real reference.

Progress in high-fidelity transfer is further hindered by the limitations of existing public datasets~\cite{Yan_2023_CVPR, li2018beautygan, jiang2020psgan, Nguyen_2021_CVPR, Gu_2019_ICCV}, which are typically low-resolution and underrepresent complex makeup styles (e.g., dense stickers, face paintings) as well as specific demographics (e.g., male portraits).
To facilitate research in high-fidelity synthesis, we introduce MakeupFaces2K (MF2K), the first 2K-resolution in-the-wild makeup portrait dataset.
Comprising 8,573 images, MF2K spans diverse makeup intensities ranging from bare skin to extreme artistic styles while maintaining broad demographic diversity.
Based on this dataset, we construct training triplets and derive a curated evaluation set that specifically emphasizes complex, fine-grained patterns and cross-domain transfers.

In summary, our main contributions are:
\begin{itemize}
    % \vspace{-2pt}
    \item \textbf{Reality-Anchored Refinement.}
    We propose ART, a two-stage DiT framework that breaks the pseudo-target ceiling.
    By anchoring a refinement cycle to the real reference, our model effectively overrides pseudo-target artifacts while stabilizing training via a controlled-noise bottleneck.
    \item \textbf{A 2K-Resolution In-the-Wild Makeup Dataset.}
    We introduce MakeupFaces2K (MF2K), the first 2K-resolution makeup portrait dataset with 8,573 images, covering diverse makeup intensities and underrepresented demographics to support high-fidelity makeup transfer and related tasks.
    \item \textbf{State-of-the-Art Performance.}
    Our method achieves superior makeup fidelity across diverse styles, while consistently preserving fine-grained details and source identity, even under occlusions and extreme expressions.
\end{itemize}

\section{Related Work}
\label{sec:related}

\subsection{Unpaired and Weakly Supervised Transfer}
Due to the scarcity of paired data, early frameworks formulated makeup transfer strictly as an unpaired image-to-image translation task.
Driven primarily by GANs~\cite{goodfellow2020generative}, methods like CycleGAN~\cite{zhu2017unpaired}, LADN~\cite{Gu_2019_ICCV}, and CSD-MT~\cite{sun2024content} relied on weak priors, such as domain adversarial losses, to constrain the cosmetic mapping.
To better handle localized cosmetics, subsequent approaches introduced spatial alignment and style-code controls for semantic correspondences~\cite{kips2020gan, deng2021spatially, sun2022ssat, sun2023ssat++}.
Recently, diffusion models~\cite{ho2020denoising, song2020denoising, rombach2022high, peebles2023scalable} have emerged as powerful generative backbones.
Building on this, SHMT~\cite{sun2024shmt} proposes a self-supervised hierarchical strategy to decompose and progressively inject makeup textures, while MAD~\cite{ruan2025mad} treats makeup tasks as cross-domain translations governed by unified domain embeddings.
Despite leveraging powerful generative priors, these paradigms lack a ground-truth transfer target and can therefore supervise the output only indirectly.
When dense spatial correspondence is ambiguous, they often struggle to achieve precise placement and recover fine-grained details (e.g., dense stickers, glitter), exposing the inherent limitation of relying solely on weak unpaired supervision.

\subsection{Surrogate Supervision via Pseudo-Targets}

To inject stronger supervisory signals, a dominant line of research constructs synthetic paired data as surrogate ground-truth.
Many methods enforce color and tone consistency via histogram matching~\cite{li2018beautygan, jiang2020psgan, liu2021psgan++, xiang2022ramgan, Yan_2023_CVPR}, or explicitly model structured makeup patterns~\cite{Nguyen_2021_CVPR}.
To further align spatial placement, other works construct pseudo-targets via geometric warping~\cite{chang2018pairedcyclegan, wan2022facial, yang2022elegant} or derive feature-level semantic targets~\cite{zhu2022semi}.

Driven by large-scale foundation models, state-of-the-art diffusion frameworks now heavily rely on external editing engines to synthesize massive pseudo-paired datasets~\cite{zhang2025stablemakeup, wu2025evomakeup, he2025beautydiffusion}.
StableMakeup~\cite{zhang2025stablemakeup} and EvoMakeup~\cite{wu2025evomakeup} construct synthetic pseudo-targets via dedicated data-construction pipelines, directly regressing the model outputs to these generated targets.
While such regression provides essential spatial guidance and improves global layout, minimizing deviation from these imperfect pseudo-targets inevitably propagates their biases and artifacts into the prediction.
This pseudo-target ceiling motivates us to seek supervision beyond synthetic approximations.

\subsection{Cycle Consistency in Generative Modeling}
Cycle consistency~\cite{zhu2017unpaired} is a foundational mechanism widely adopted to regularize domain invertibility.
Recently, this concept has also been retrofitted into diffusion pipelines (e.g., CycleNet~\cite{xu2023cyclenet}, CycleDiff~\cite{zou2025cyclediff}) to stabilize unpaired manipulation and zero-shot editing.

However, existing cycle objectives primarily verify source recoverability, acting as domain-level regularizers rather than supervising the forward generation with instance-level detail.
Consequently, while they improve overall translation stability, they remain insufficient for the precise, fine-grained texture recovery that complex cosmetics demand.

In contrast, our reality-anchored refinement cycle reconstructs the real reference itself, verifying whether the transfer prediction carries reference-specific makeup cues.
Rather than imposing a domain-level invertibility penalty, it provides a corrective, instance-level supervisory signal that drives the network to preserve fine-grained cosmetic details and override pseudo-target artifacts, bridging weak domain regularization and exact textural grounding.

\section{Methodology}
\label{sec:method}

In this section, we first introduce the preliminaries (Sec.~\ref{sec:preliminaries}), then present the two-stage strategy with a focus on the reality-anchored refinement cycle (Sec.~\ref{sec:training}), and finally describe our high-resolution makeup dataset (Sec.~\ref{sec:data}).

\subsection{Preliminaries}
\label{sec:preliminaries}
Given a source image $I_{\mathrm{src}}$ and a reference image $I_{\mathrm{ref}}$, the goal of makeup transfer is to synthesize an output that preserves the identity and geometry of $I_{\mathrm{src}}$ while faithfully adopting the cosmetic style of $I_{\mathrm{ref}}$.
Since paired ground-truth is unavailable, we construct a synthetic pseudo-target $I_{\mathrm{pseudo}}$ and extract a bare-skin counterpart $I_{\mathrm{bare\_ref}}$ of the reference via an auxiliary makeup remover for our refinement cycle.
We operate in the latent space of a pretrained VAE~\cite{kingma2013auto}, denoting the latent representation of any image $I$ as $z = \mathcal{E}(I)$.

We train the conditional DiT using the Flow Matching (FM) objective~\cite{lipman2022flow}.
For a target latent $z$ and condition $c$, we sample noise $\epsilon \sim \mathcal{N}(0,I)$ and a timestep $\sigma \in [0,1]$ to construct the noisy latent $z_{\sigma} = (1-\sigma)z + \sigma \epsilon$.
The model predicts a velocity field $v_{\theta}(z_{\sigma},\sigma; c)$ and the objective is defined as:
\begin{equation}
    \mathcal{L}_{\mathrm{FM}}(\theta; z, c) = \mathbb{E}_{\epsilon,\,\sigma} \Big[ w(\sigma) \big\| v_{\theta}(z_{\sigma},\sigma;c) - (\epsilon - z) \big\|_2^2 \Big]\,,
    \label{eq:fm_loss}
\end{equation}
where $w(\sigma)$ is a standard weighting function.

Following standard DiT conventions for image editing, condition injection is implemented via token concatenation~\cite{tan2025ominicontrol, tan2025ominicontrol2, zhang2025easycontrol}.
The input sequence to the transformer is constructed by concatenating the position-encoded tokens of the noisy target, the source, and the reference latents.

\begin{figure}[!t]
    \centering
    \includegraphics[width=\linewidth]{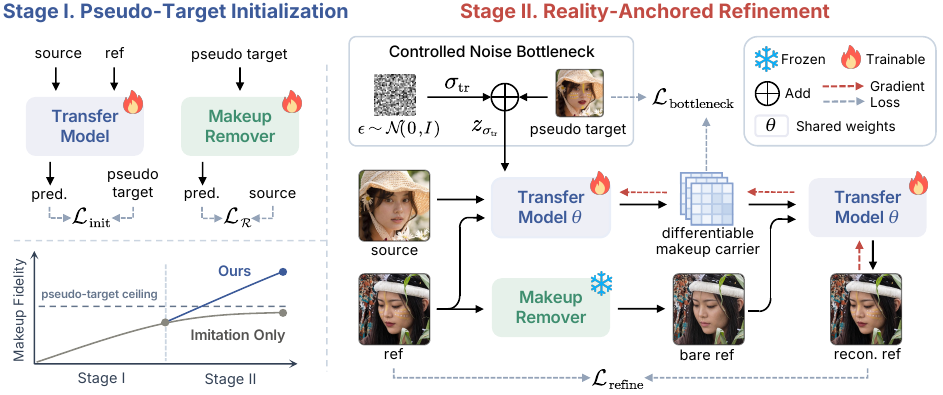}
    \caption{\textbf{Overview of ART.}
        ART is a two-stage framework that shifts supervision from the synthetic pseudo-target to the real reference.
        \textbf{Stage~I} performs pseudo-target initialization, training the transfer model for global makeup placement and an auxiliary makeup remover for bare-skin extraction.
        \textbf{Stage~II} predicts a differentiable makeup carrier $\hat{z}$ at noise level $\sigma_{\mathrm{tr}}$, then reconstructs the real reference from its bare-skin counterpart conditioned on $\hat{z}$; as $\hat{z}$ stays differentiable, the gradient of $\mathcal{L}_{\mathrm{refine}}$ back-propagates into the transfer prediction to recover fine-grained makeup cues and override synthetic artifacts, while $\mathcal{L}_{\mathrm{bottleneck}}$ preserves the global structure.}
    \label{fig:pipeline}
\end{figure}

\subsection{Two-Stage Training Strategy}
\label{sec:training}

To break the pseudo-target ceiling imposed by $I_{\mathrm{pseudo}}$, which is synthesized by a large-scale image editing model, we propose a two-stage training strategy that shifts the learning paradigm toward differentiable reality-anchored reconstruction.
An overview is shown in Fig.~\ref{fig:pipeline}.

{\textbf{Stage I: Pseudo-Target Initialization.}}
In this stage, we establish the foundation for the subsequent refinement cycle by independently training the main transfer model and an auxiliary makeup remover $\mathcal{R}$.

\emph{Transfer Model Initialization.}
The transfer model is initialized to imitate the synthetic pseudo-target $I_{\mathrm{pseudo}}$.
We minimize the standard FM objective conditioned on the source and reference:
\begin{equation}
    \mathcal{L}_{\mathrm{init}} = \mathcal{L}_{\mathrm{FM}}\big(\theta; z_{\mathrm{pseudo}}, z_{\mathrm{src}}, z_{\mathrm{ref}}\big).
    \label{eq:stage1_loss}
\end{equation}
This establishes basic semantic alignment and global makeup placement.
However, its performance is inherently bounded by the artifacts (e.g., incorrect details or identity drift) present in $I_{\mathrm{pseudo}}$.

\emph{Bare-Skin Counterpart Extraction.}
In parallel, we train an auxiliary makeup remover $\mathcal{R}$ to recover a bare-skin face from its makeup counterpart, isolating the facial identity to anchor the Stage~II refinement cycle.
We implement $\mathcal{R}$ as a one-step denoising model and supervise it with the pair $(I_{\mathrm{pseudo}}, I_{\mathrm{src}})$, where $I_{\mathrm{pseudo}}$ is the source $I_{\mathrm{src}}$ rendered with the reference makeup.
Its optimization is driven by a standard FM objective, complemented by auxiliary regularizers for perceptual fidelity, identity preservation, and structural consistency:
\begin{equation}
\mathcal{L}_{\mathcal{R}} = \mathcal{L}_{\mathrm{FM}}\big(\phi; z_{\mathrm{src}}, z_{\mathrm{pseudo}}\big) + \lambda_{\mathrm{lpips}}\mathcal{L}_{\mathrm{lpips}} + \lambda_{\mathrm{id}}\mathcal{L}_{\mathrm{id}} + \lambda_{\mathrm{lmk}}\mathcal{L}_{\mathrm{lmk}}\,.
\label{eq:remover_loss}
\end{equation}
where $\mathcal{L}_{\mathrm{lpips}}$~\cite{zhang2018unreasonable} improves perceptual quality, $\mathcal{L}_{\mathrm{id}}$ enforces identity consistency via ArcFace embeddings~\cite{deng2019arcface}, and $\mathcal{L}_{\mathrm{lmk}}$~\cite{bulat2017far} preserves the facial geometric structure.
Once trained, $\mathcal{R}$ is applied to the reference to obtain its bare-skin counterpart, $I_{\mathrm{bare\_ref}} = \mathcal{R}(I_{\mathrm{ref}})$.

{\textbf{Stage II: Reality-Anchored Refinement.}}
This stage shifts supervision from the synthetic pseudo-target to the real reference, anchoring the model to authentic makeup detail rather than a synthetic approximation.

Starting from the pseudo-target latent $z_{\mathrm{pseudo}}$, we perturb it with Gaussian noise at noise level $\sigma_{\mathrm{tr}}$ to obtain the noisy latent $z_{\sigma_{\mathrm{tr}}}$.
A one-step Euler update is then applied to predict the makeup carrier $\hat{z}$:
\begin{equation}
    \hat{z} = z_{\sigma_{\mathrm{tr}}} - \sigma_{\mathrm{tr}}\, v_{\theta}\!\left(z_{\sigma_{\mathrm{tr}}},\sigma_{\mathrm{tr}};\,z_{\mathrm{src}},z_{\mathrm{ref}}\right)\,.
    \label{eq:stage2_transfer}
\end{equation}
Rather than decoding $\hat{z}$ into a pixel-space image, we retain it within the continuous computational graph as a differentiable \textit{makeup carrier}.

\emph{Reality-Anchored Reconstruction.}
The same transfer model (with shared weights) is then repurposed to reconstruct the real reference latent $z_{\mathrm{ref}}$, conditioned solely on the bare-skin counterpart $z_{\mathrm{bare\_ref}}$ and the predicted carrier $\hat{z}$.
The corresponding refinement loss $\mathcal{L}_{\mathrm{refine}}$ is defined as:
\begin{equation}
    \mathcal{L}_{\mathrm{refine}} = \mathcal{L}_{\mathrm{FM}}\big(\theta; z_{\mathrm{ref}}, z_{\mathrm{bare\_ref}}, \hat{z}\big)\,.
    \label{eq:stage2_restore}
\end{equation}
Notably, since the makeup carrier $\hat{z}$ remains differentiable, the gradient of $\mathcal{L}_{\mathrm{refine}}$ back-propagates smoothly through the DiT blocks directly into the velocity field $v_{\theta}$ responsible for the initial transfer prediction (Eq.~\ref{eq:stage2_transfer}).
This formulation explicitly penalizes the omission of fine-grained makeup cues, encouraging the transfer prediction to preserve the details necessary for faithfully reconstructing the real reference.

While $\mathcal{L}_{\mathrm{refine}}$ provides a high-fidelity refinement constraint, optimizing it alone could lead to degenerate solutions, such as neglecting the source identity to trivialize the refinement objective (Suppl.~Sec.~\ref{subsec:ablation_bottleneck_loss}).
To mitigate this and stabilize the global makeup placement, we introduce $\mathcal{L}_{\mathrm{bottleneck}}$ as a structural regularizer.
Specifically, it enforces consistency with the pseudo-target prior exclusively at the noise bottleneck $\sigma_{\mathrm{tr}}$:
\begin{equation}
    \mathcal{L}_{\mathrm{bottleneck}} = \mathbb{E}_{\epsilon} \Big[ w(\sigma_{\mathrm{tr}})\, \big\| v_{\theta}\!\left(z_{\sigma_{\mathrm{tr}}},\sigma_{\mathrm{tr}};\, z_{\mathrm{src}},z_{\mathrm{ref}}\right) - (\epsilon-z_{\mathrm{pseudo}}) \big\|_2^2 \Big]\,,
    \label{eq:bottleneck_loss}
\end{equation}
where $z_{\sigma_{\mathrm{tr}}}$ is the noised latent defined in Eq.~\eqref{eq:stage2_transfer}.

The overall objective for Stage~II is a weighted combination of the refinement loss $\mathcal{L}_{\mathrm{refine}}$ and the structural regularizer $\mathcal{L}_{\mathrm{bottleneck}}$:
\begin{equation}
    \mathcal{L}_{\mathrm{s2}} = \lambda_{\mathrm{refine}}\,\mathcal{L}_{\mathrm{refine}} + \lambda_{\mathrm{bot}}\,\mathcal{L}_{\mathrm{bottleneck}}\,.
    \label{eq:stage2_objective}
\end{equation}
Here, the controlled noise level $\sigma_{\mathrm{tr}}$ serves as a semantic information bottleneck that modulates the generative trade-off.
$\mathcal{L}_{\mathrm{bottleneck}}$ preserves the coarse makeup placement as a structural foundation, while affording the model sufficient flexibility to recover high-fidelity texture via $\mathcal{L}_{\mathrm{refine}}$.
An appropriate $\sigma_{\mathrm{tr}}$ (ablated in Sec.~\ref{sec:ablation}) ensures that global structure is robustly preserved while overriding synthetic artifacts.

\subsection{2K-Resolution Makeup Dataset}
\label{sec:data}

\begin{figure}[!t]
    \centering
    \includegraphics[width=0.97\linewidth]{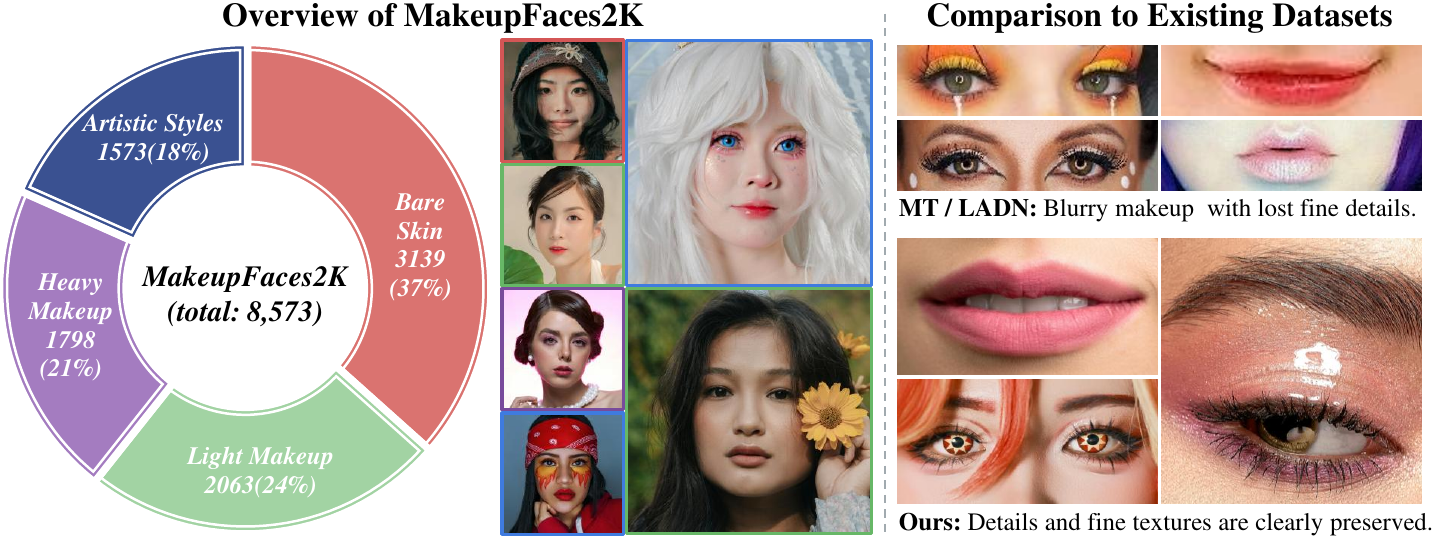}
    \caption{\textbf{MF2K, the first 2K-resolution in-the-wild makeup dataset.}
        MF2K contains 8,573 images spanning four makeup intensities from bare skin to artistic styles, with broad demographic diversity.
        Compared with existing datasets, it captures fine-grained makeup textures and thus serves as a demanding benchmark for high-fidelity transfer.
        Zoom in to see the fine details.}
    \label{fig:dataset_card}
\end{figure}

\paragraph{\textbf{MF2K.}}
Existing makeup transfer datasets are often limited in image resolution and provide insufficient coverage of challenging cases~\cite{Yan_2023_CVPR, li2018beautygan, jiang2020psgan, Nguyen_2021_CVPR, Gu_2019_ICCV}, which fundamentally hinders the learning and evaluation of localized, fine-grained details.
To address these limitations, we introduce MakeupFaces2K (MF2K), the first 2K-resolution makeup portrait dataset comprising 8,573 images at $2048\times2048$ resolution.
MF2K covers a wide range of makeup intensities, including bare skin (3,139), light makeup (2,063), heavy makeup (1,798), and artistic styles (1,573), and includes diverse demographics with improved coverage of male portraits.
With its high-resolution and diverse styles, MF2K not only enables high-fidelity training but also constitutes a demanding benchmark for fine-grained makeup transfer.
We summarize the dataset statistics and compare MF2K with existing public datasets in Fig.~\ref{fig:dataset_card}.

{\textbf{Pseudo-target triplets.}}
As the MF2K dataset contains only bare-skin and makeup portraits and lacks paired ground-truth for makeup transfer, we construct the training triplets by sampling source–reference pairs $(I_{\mathrm{src}}, I_{\mathrm{ref}})$ and generating a corresponding pseudo-target $I_{\mathrm{pseudo}}$ using a large-scale image editing model $\mathcal{G}$:
\begin{equation}
    I_{\mathrm{pseudo}} = \mathcal{G}(I_{\mathrm{src}}, I_{\mathrm{ref}}).
    \label{eq:pseudo_target}
\end{equation}
To ensure balanced style coverage, we perform stratified sampling across makeup categories, yielding 2,000 source–reference pairs and their corresponding training triplets $\big(I_{\mathrm{src}}, I_{\mathrm{ref}}, I_{\mathrm{pseudo}}\big)$.
For the remaining images, which lack a pseudo-target, we predict the carrier from the source latent $z_{\mathrm{src}}$ in place of $z_{\mathrm{pseudo}}$, and train the transfer model with the refinement loss alone, setting $\lambda_{\mathrm{bot}}$ to 0.
More details about dataset statistics and sampling ratios are provided in Suppl.~Sec.~\ref{subsec:mf2k_details}.

\section{Experiments}
\label{sec:experiments}

\subsection{Experimental Settings}
\label{sec:exp_settings}

\textbf{Implementation details.}
We employ the pre-trained FLUX.1-Kontext-dev~\cite{batifol2025flux} as our backbone and extend its Rotary Position Embedding (RoPE) with an explicit image-level index to support multi-image inputs.
We fine-tune it via Low-Rank Adaptation (LoRA)~\cite{hu2022lora} using the proposed two-stage training strategy.
The LoRA rank and scaling factor are both set to 32.
We optimize the model using Prodigy~\cite{mishchenko2023prodigy} with a learning rate of 1.
Training is conducted on 4 NVIDIA H20 GPUs, with a total batch size of 16 for 512$\times$512 inputs and 4 for 2K-resolution inputs.
Stage I is trained for 10k steps as pseudo-target initialization, followed by 10k steps of reality-anchored refinement in Stage II.
During Stage II, the controlled noise level is fixed at $\sigma_{\mathrm{tr}}=0.6$ with loss weights $\lambda_{\mathrm{refine}}=1$ and $\lambda_{\mathrm{bot}}=0.2$.
The remover $\mathcal{R}$ is pre-trained separately and remains frozen during the refinement stage.
Other parameters are detailed in Suppl.~Sec.~\ref{subsec:implementation_details}.

\noindent\textbf{Baselines.}
We compare our method with representative approaches across three categories: GAN-based makeup transfer models, including PSGAN~\cite{jiang2020psgan} and EleGANt~\cite{yang2022elegant}; diffusion-based methods, including StableMakeup~\cite{zhang2025stablemakeup}, SHMT~\cite{sun2024shmt}, and MAD~\cite{ruan2025mad}; and leading commercial image editing models, including Nano~Banana~Pro~\cite{deepmind_gemini_image_pro} and GPT~Image~1.5~\cite{openai_gpt_image_15_model}.
For all open-source baselines, we use the official implementations and follow the recommended settings.

\subsection{Evaluation}
\label{sec:exp_data_eval}

{\textbf{Datasets.}}
We evaluate our method on three public makeup datasets including MT~\cite{li2018beautygan}, MT-Wild~\cite{jiang2020psgan}, and LADN~\cite{Gu_2019_ICCV}, as well as our MF2K.
For each dataset, we randomly sample 100 source--reference pairs with non-overlapping identities to form the evaluation set.
In particular, to assess performance under complex scenarios, we sample the reference images in the MF2K evaluation set exclusively from its artistic-makeup subset.
We enforce a strict separation between training and evaluation splits across all datasets to prevent data leakage.

{\textbf{Metrics.}}
We evaluate makeup transfer quality from four dimensions: makeup similarity, identity preservation, background stability, and image quality.
Makeup similarity is measured using a rubric-based Visual Language Model (VLM) evaluation, which assesses how accurately makeup textures and patterns from the reference are transferred to the output.
Concretely, we use Gemini~3~Pro~\cite{deepmind_gemini_3_pro} and Qwen2.5-VL-72B-Instruct~\cite{bai2025qwen25vltechnicalreport} as judges to assign a makeup similarity score on a 0--10 scale, denoted as MSim$_G$ and MSim$_Q$, respectively.
For fairness, all images fed into the VLM judges are resized to $512\times512$.
Identity preservation is quantified by an ArcFace-based similarity metric (ID)~\cite{deng2019arcface}, computed as the cosine similarity between ArcFace embeddings of the generated image and the source image.
Background stability is evaluated using L2-M~\cite{zhang2025stablemakeup}, which computes the pixel-averaged $L_2$ distance within the unedited background mask.
We assess image quality with Fr\'echet Inception Distance (FID)~\cite{heusel2017gans}, computed between generated outputs and real images from the evaluation set.
Although prior work often employs feature-based distances (e.g., CLIP~\cite{radford2021learning} and DINO~\cite{oquab2023dinov2, simeoni2025dinov3}) for makeup similarity, we find that these global metrics can be unreliable under certain makeup conditions.
We analyze the failure modes and report the corresponding results together with qualitative evidence in Suppl.~Sec.~\ref{subsec:metric_failures}.

\subsection{Experiment Results}
\label{sec:exp_results}

\begin{figure}[!t]
    \centering
    \includegraphics[width=0.97\linewidth]{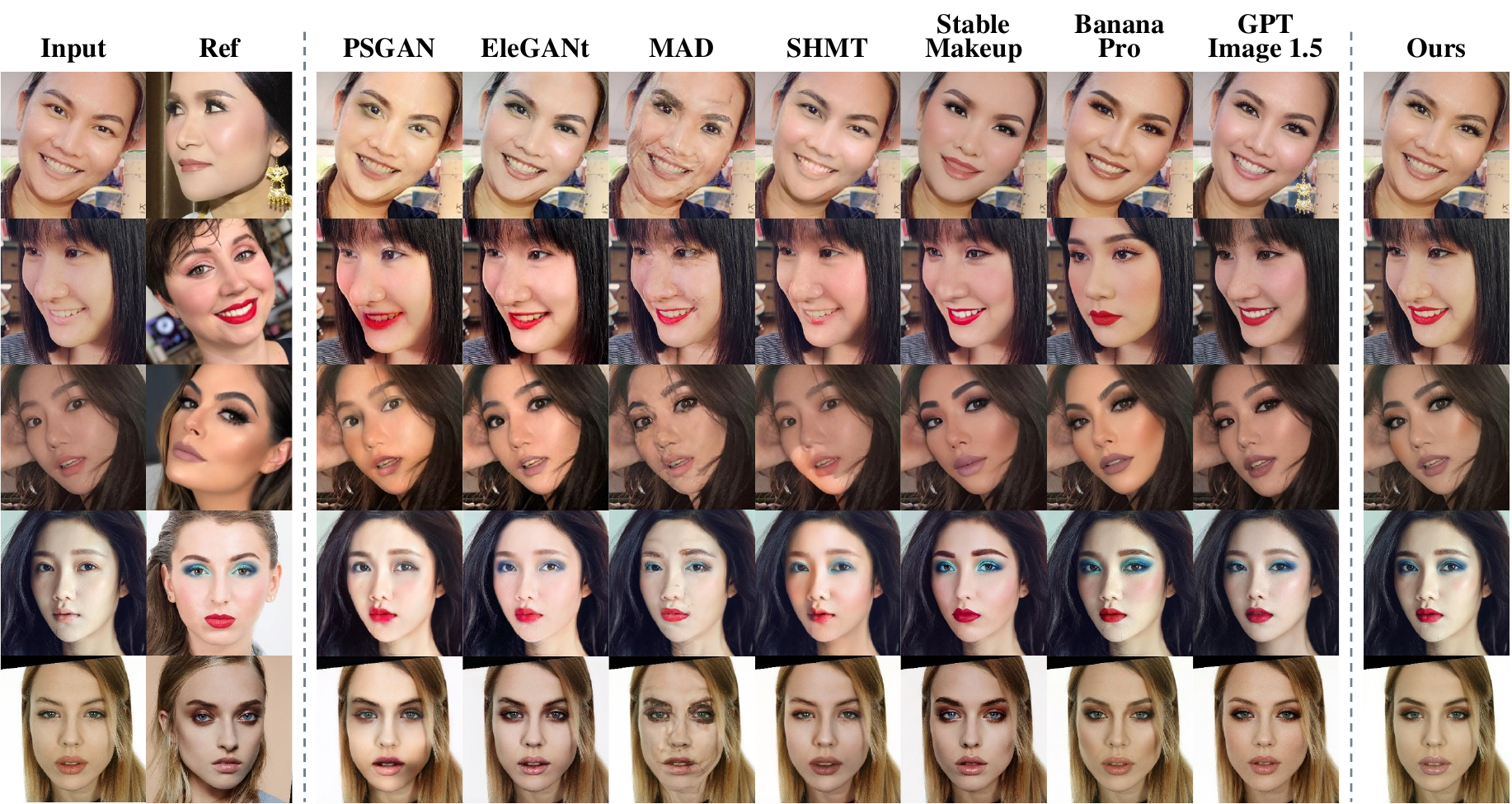}
    \caption{\textbf{Qualitative comparison on simple makeup.}
        Zoom in to see details.}
    \label{fig:qualitative_easy}
    % \vspace{2pt}
\end{figure}

\begin{figure}[!t]
    \centering
    \includegraphics[width=0.97\linewidth]{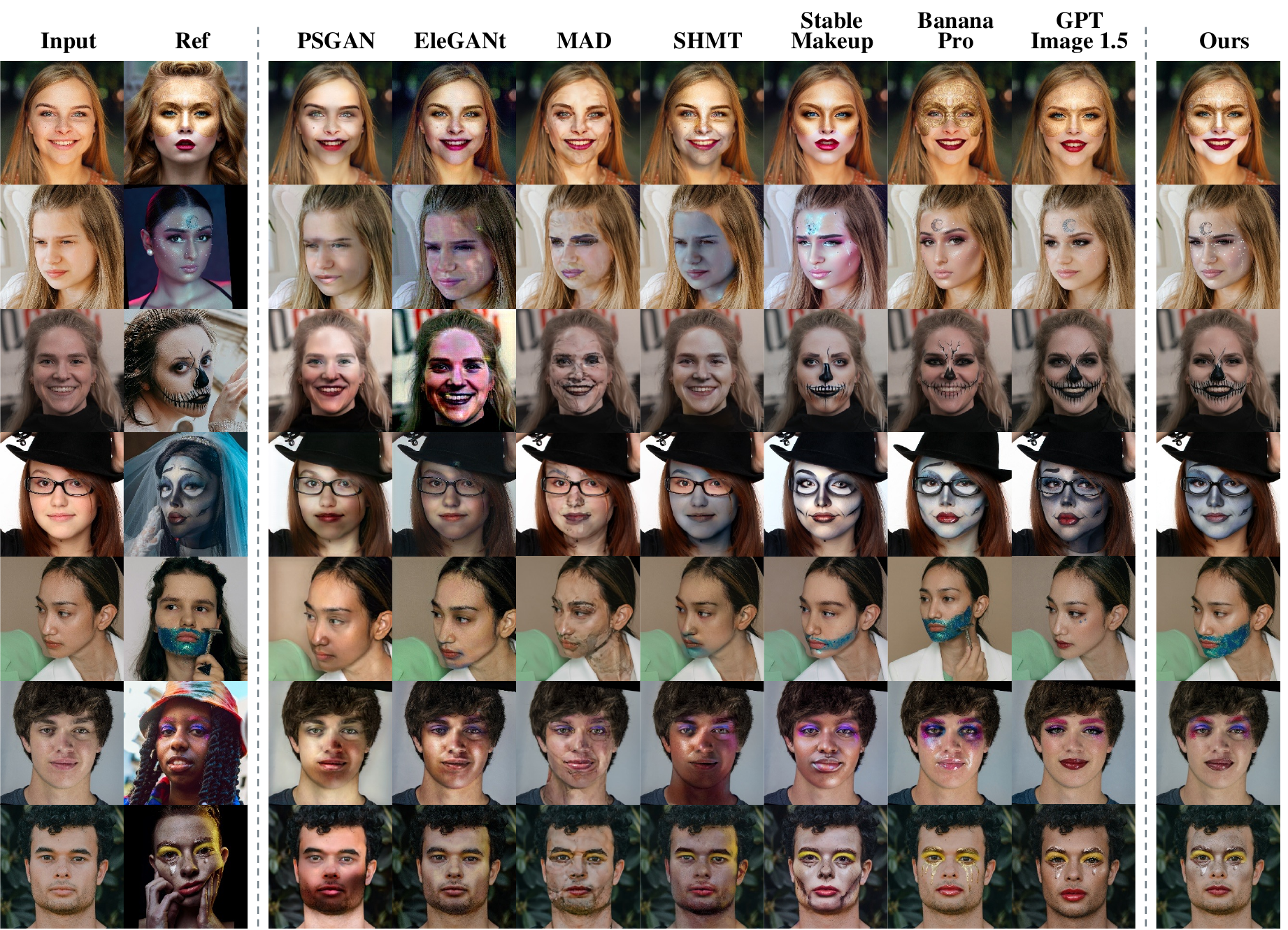}
    \caption{\textbf{Qualitative comparison on complex makeup.}
        Zoom in to see details.}
    \label{fig:qualitative_hard}
    % \vspace{-1pt}
\end{figure}

{\textbf{Qualitative results.}}
Fig.~\ref{fig:qualitative_easy} and Fig.~\ref{fig:qualitative_hard} compare performance on simple daily styles and complex cosmetics (e.g., dense stickers, face paintings).
GAN-based methods (PSGAN, EleGANt) manage basic tones but produce unnatural blending on simple styles, and fail completely on complex geometries by merely applying global color tints.
Previous diffusion methods (SHMT, StableMakeup) attempt to generate fine-grained patterns but often suffer from severe blurring, skin tone shifts, and structural misregistration.
MAD exhibits severe facial distortions, while commercial editing models frequently alter the intrinsic facial geometry and identity.
In contrast, by leveraging the proposed reality-anchored refinement cycle, our method faithfully transfers the reference cosmetics while preserving sharp boundaries and source identity in both simple and complex settings.

\begin{table*}[t]
    \centering
    \caption{\textbf{Quantitative comparison on four makeup transfer evaluation sets.}
        Banana Pro and GPT 1.5 denote Nano~Banana~Pro and GPT~Image~1.5.
        Best and second-best are in \textbf{bold} and \underline{underline}.}
    \scriptsize
    \setlength{\tabcolsep}{1.8pt}
    \renewcommand{\arraystretch}{1.08}
    \begin{tabular}{l|ccccc|ccccc}
        \toprule
        & \multicolumn{5}{c|}{\textbf{MT Test Set~\cite{li2018beautygan}}} & \multicolumn{5}{c}{\textbf{LADN Test Set~\cite{Gu_2019_ICCV}}} \\
        \cmidrule(lr){2-6}\cmidrule(lr){7-11}
        \multirow{-2.6}{*}{\makecell[c]{\textbf{Methods}}}  &
        MSim$_G$$\uparrow$ &
        MSim$_Q$$\uparrow$ &
        ID$\uparrow$ &
        L2-M$\downarrow$ &
        FID$\downarrow$
        &
        MSim$_G$$\uparrow$ &
        MSim$_Q$$\uparrow$ &
        ID$\uparrow$ &
        L2-M$\downarrow$ &
        FID$\downarrow$\\
        \midrule
        PSGAN~\cite{jiang2020psgan}                & 6.13 & 5.26 & .80 & 31.06 & 98.60 & 3.79 & 3.44 & .80 & 17.32 & 108.21 \\
        EleGANt~\cite{yang2022elegant}            & 6.98 & 4.94 & .83 & 34.95 & 95.56 & 5.64 & 4.82 & \underline{.81} & 20.34 & 90.52 \\
        MAD~\cite{ruan2025mad}                    & 6.60 & 5.10 & .69 & -     & 98.77 & 3.78 & 3.49 & .60 & -     & 96.02 \\
        SHMT~\cite{sun2024shmt}                   & 5.87 & 4.22 & \underline{.86} & \underline{6.34} & 102.57 & 4.78 & 5.00 & \textbf{.85} & \underline{5.57} & 91.58 \\
        StableMakeup~\cite{zhang2025stablemakeup} & 8.34 & \underline{7.30} & .49 & 9.55  & \textbf{82.70} & 7.81 & 7.89 & .47 & 8.91  & \textbf{65.99} \\
        Banana~Pro~\cite{deepmind_gemini_image_pro} & 8.34 & 7.07 & .67 & 9.98 & 86.66 & 8.18 & 7.71 & .67 & 38.40 & 70.75 \\
        GPT~1.5~\cite{openai_gpt_image_15_model}   & \underline{8.69} & \textbf{7.51} & .36 & 33.22 & \underline{85.94} & \underline{8.85} & \textbf{8.13} & .35 & 43.59 & 69.14 \\
        \textbf{Ours}                             & \textbf{8.96} & 7.12 & \textbf{.87} & \textbf{3.94} & 87.72 & \textbf{9.14} & \underline{8.04} & \underline{.81} & \textbf{3.54} & \underline{68.67} \\
        \midrule
        & \multicolumn{5}{c|}{\textbf{MT-Wild Test Set~\cite{jiang2020psgan}}} & \multicolumn{5}{c}{\textbf{Our MF2K Test Set}} \\
        \cmidrule(lr){2-6}\cmidrule(lr){7-11}
        \multirow{-2.6}{*}{\makecell[c]{\textbf{Methods}}}  & MSim$_G$$\uparrow$ & MSim$_Q$$\uparrow$ & ID$\uparrow$ & L2-M$\downarrow$ & FID$\downarrow$
        & MSim$_G$$\uparrow$ & MSim$_Q$$\uparrow$ & ID$\uparrow$ & L2-M$\downarrow$ & FID$\downarrow$ \\
        \midrule
        PSGAN~\cite{jiang2020psgan}                & 3.65 & 2.71 & .82 & 42.22 & 112.61 & 1.49 & 0.74 & \underline{.79} & 45.23 & 166.79 \\
        EleGANt~\cite{yang2022elegant}            & 5.30 & 4.30 & .82 & 43.86 & 113.24 & 2.60 & 1.87 & \underline{.79} & 48.88 & 139.07 \\
        MAD~\cite{ruan2025mad}                    & 3.93 & 2.85 & .61 & -     & 118.43 & 1.86 & 1.37 & .57 & -     & 156.41 \\
        SHMT~\cite{sun2024shmt}                   & 3.11 & 2.45 & \textbf{.88} & \underline{8.44} & 115.81 & 3.47 & 2.50 & \textbf{.82} & \underline{6.71} & 133.97 \\
        StableMakeup~\cite{zhang2025stablemakeup} & 8.21 & \textbf{7.62} & .48 & 9.91 & 86.51 & 6.73 & 7.04 & .43 & 9.26 & \textbf{100.60} \\
        Banana~Pro~\cite{deepmind_gemini_image_pro} & 8.45 & 7.39 & .53 & 10.38 & \textbf{80.48} & 8.34 & 7.06 & .65 & 13.27 & 104.45 \\
        GPT~1.5~\cite{openai_gpt_image_15_model}   & \underline{8.74} & \underline{7.53} & .28 & 35.27 & \underline{84.07} & \underline{8.43} & \textbf{7.69} & .35 & 28.32 & \underline{102.94} \\
        \textbf{Ours}                             & \textbf{8.77} & 7.00 & \underline{.85} & \textbf{4.37} & 91.11 & \textbf{9.22} & \underline{7.68} & .74 & \textbf{4.31} & 103.27 \\
        \bottomrule
    \end{tabular}
    % \vspace{-8pt}
    \label{tab:main}
\end{table*}

\noindent{\textbf{Quantitative results.}}
Tab.~\ref{tab:main} shows that our method achieves the best overall balance across the four evaluation sets.
It ranks first on MSim$_G$ across all four sets, by the largest margin on the artistic MF2K set (9.22 vs.\ 8.43), and remains competitive on MSim$_Q$; we provide results from more VLM judges in Suppl.~Sec.~\ref{subsec:metric_failures}.
For identity, ART obtains the highest ID on MT and the second-highest on LADN and MT-Wild, whereas methods that achieve higher identity, such as SHMT, do so at the expense of makeup fidelity.
Background stability is likewise the strongest, with the lowest L2-M on every set, suppressing the unintended edits commonly inherited from synthetic pseudo-targets.
On FID, our method scores higher than StableMakeup.
FID measures global distributional similarity and is known to diverge from fine-grained perceptual quality~\cite{jayasumana2024rethinking,stein2023exposing}; in our case, faithfully reproducing the reference's intricate artistic textures introduces patterns that are out-of-domain relative to the average-face statistics encoded by FID.
Overall, these results demonstrate that the reality-anchored refinement cycle improves makeup fidelity without incurring the identity drift or background instability exhibited by competing methods.

\begin{wraptable}{r}{0.56\textwidth}
    \centering
    \vspace{-8pt}
    \caption{\textbf{User study results.}}
    \label{tab:user_study}
    \scriptsize
    \setlength{\tabcolsep}{2pt}
    \renewcommand{\arraystretch}{1.08}
    \begin{tabular}{l|ccc}
        \toprule
        \textbf{Methods} & \makecell[c]{Makeup \\Similarity}$\uparrow$ & \makecell[c]{ID \\Consistency}$\uparrow$ & \makecell[c]{Image \\Quality}$\uparrow$ \\
        \midrule
        PSGAN~\cite{jiang2020psgan}                 & 1.61 & 3.00 & 2.07 \\
        EleGANt~\cite{yang2022elegant}             & 2.01 & 3.03 & 2.23 \\
        MAD~\cite{ruan2025mad}                     & 1.51 & 1.97 & 1.51 \\
        SHMT~\cite{sun2024shmt}                    & 1.73 & \underline{3.21} & 2.19 \\
        StableMakeup~\cite{zhang2025stablemakeup}  & 2.75 & 2.28 & 2.73 \\
        Banana Pro~\cite{deepmind_gemini_image_pro} & \underline{3.43} & 3.12 & 3.81 \\
        GPT 1.5~\cite{openai_gpt_image_15_model}   & 3.31 & 2.15 & \underline{3.82} \\
        \textbf{Ours}                               & \textbf{3.67} & \textbf{4.03} & \textbf{3.83} \\
        \bottomrule
    \end{tabular}
    \vspace{-18pt}
\end{wraptable}

\noindent\textbf{User Study.}
\label{sec:user_study}
To assess perceptual preferences, we collected 864 ratings from 21 participants, who scored the anonymized results in randomized order on a 1--5 Likert scale for Makeup Similarity, ID Consistency, and Image Quality; participant demographics and the rating rubrics are detailed in Suppl.~Sec.~\ref{subsec:user_study_details}.
As reported in Tab.~\ref{tab:user_study}, our method ranks first on all criteria.
It achieves the highest Makeup Similarity (3.67) for faithful pattern transfer, and the best ID Consistency (4.03), indicating that improved makeup fidelity does not induce identity drift under human judgment.
Meanwhile, our Image Quality (3.83) remains on par with the strongest commercial editors.
Overall, human evaluations corroborate our quantitative findings and support the effectiveness of the reality-anchored refinement cycle.

\subsection{Ablation Studies}
\label{sec:ablation}
We conduct ablation studies to isolate the contribution of each design, covering the two-stage training strategy, the controlled-noise bottleneck, and the pseudo-target supervision and data.
Unless otherwise specified, all evaluations follow the settings detailed in Sec.~\ref{sec:exp_data_eval}.

\noindent{\textbf{Two-stage training.}}
We evaluate the effectiveness of the proposed framework by comparing the full model (Stage I \& II) against baselines trained solely via Stage~I or Stage~II.
As reported in Tab.~\ref{tab:ablation_stage}, training with Stage I alone establishes basic makeup placement but exhibits pronounced identity drift, since it merely imitates the imperfect pseudo-targets.
Conversely, training with Stage II alone preserves identity but cannot transfer makeup faithfully: without the Stage~I initialization, the makeup carrier is largely uninformative, so reconstructing the reference yields only a weak signal for learning the transfer.
Our full framework combines the complementary strengths of both stages.
It achieves the highest MSim$_{G}$ (9.22 on MF2K), capturing fine-grained cosmetic details, while substantially improving identity preservation and attaining the lowest L2-M (4.31 on MF2K) to reduce unintended background modifications.
Overall, these results validate that Stage~I is crucial for the structural foundation, while the Stage~II refinement cycle effectively breaks the pseudo-target ceiling to recover fine-grained patterns and mitigate synthetic artifacts.

\begin{table*}[t]
    \centering
    \caption{\textbf{Ablation of the two-stage training strategy.}}
    \scriptsize
    \setlength{\tabcolsep}{1.2pt}
    \renewcommand{\arraystretch}{.98}
    \begin{tabular}{l|ccccc|ccccc}
        \toprule
        & \multicolumn{5}{c|}{\textbf{MT Test Set~\cite{li2018beautygan}}} & \multicolumn{5}{c}{\textbf{LADN Test Set~\cite{Gu_2019_ICCV}}} \\
        \cmidrule(lr){2-6}\cmidrule(lr){7-11}
        \multirow{-2.3}{*}{\makecell[l]{\textbf{Variants}}} & MSim$_G$$\uparrow$ & MSim$_Q$$\uparrow$ & ID$\uparrow$ & L2-M$\downarrow$ & FID$\downarrow$
        & MSim$_G$$\uparrow$ & MSim$_Q$$\uparrow$ & ID$\uparrow$ & L2-M$\downarrow$ & FID$\downarrow$ \\
        \midrule
        Stage I only & \underline{7.81} & \textbf{7.47} & .56 & 4.87 & \textbf{84.93} & 8.30 & \underline{8.00} & .60 & 11.52 & \textbf{63.14} \\
        Stage II only & 6.96 & 5.35 & \textbf{.93} & \underline{4.45} & 92.72 & \underline{8.34} & 7.79 & \textbf{.81} & \underline{4.38} & 70.64 \\
        Stage I\&II & \textbf{8.96} & \underline{7.12} & \underline{.87} & \textbf{3.94} & \underline{87.72} & \textbf{9.14} & \textbf{8.04} & \textbf{.81} & \textbf{3.54} & \underline{68.67} \\
        \midrule
        & \multicolumn{5}{c|}{\textbf{MT-Wild Test Set~\cite{jiang2020psgan}}} & \multicolumn{5}{c}{\textbf{Our MF2K Test Set}} \\
        \cmidrule(lr){2-6}\cmidrule(lr){7-11}
        \multirow{-2.3}{*}{\makecell[l]{\textbf{Variants}}} & MSim$_G$$\uparrow$ & MSim$_Q$$\uparrow$ & ID$\uparrow$ & L2-M$\downarrow$ & FID$\downarrow$
        & MSim$_G$$\uparrow$ & MSim$_Q$$\uparrow$ & ID$\uparrow$ & L2-M$\downarrow$ & FID$\downarrow$ \\
        \midrule
        Stage I only & \textbf{8.77} & \textbf{7.78} & .46 & \underline{4.59} & \textbf{83.73} & \underline{8.82} & \textbf{7.74} & .57 & 7.68 & \textbf{96.85} \\
        Stage II only & 6.60 & 5.50 & \textbf{.91} & 5.02 & \underline{101.58} & 8.08 & 6.72 & \underline{.70} & \underline{4.98} & \underline{103.97} \\
        Stage I\&II & \textbf{8.77} & \underline{7.00} & \underline{.85} & \textbf{4.37} & 91.11 & \textbf{9.22} & \underline{7.68} & \textbf{.74} & \textbf{4.31} & 103.27 \\
        \bottomrule
    \end{tabular}
    \label{tab:ablation_stage}
    % \vspace{-10pt}
\end{table*}

\begin{figure}[!t]
    \centering
    \includegraphics[width=\linewidth]{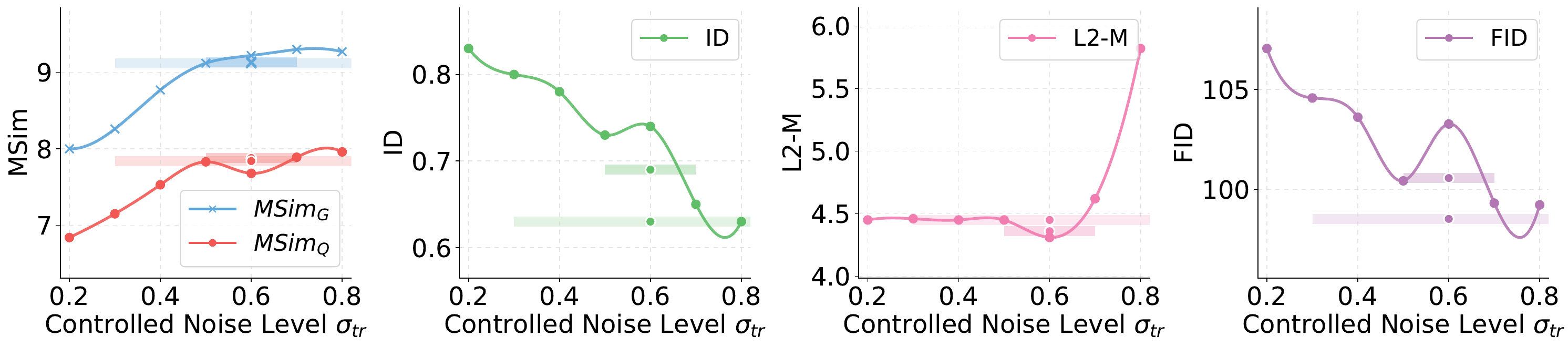}
    \caption{\textbf{Sensitivity to the controlled noise level $\sigma_{\mathrm{tr}}$.}
        Markers denote discrete, fixed $\sigma_{\mathrm{tr}}$ values, and shaded horizontal bands denote $\sigma_{\mathrm{tr}}$ sampled from the corresponding continuous interval.}
    \label{fig:sigma_figure}
\end{figure}

\noindent{\textbf{Controlled-noise bottleneck.}}
We analyze the sensitivity of the controlled noise level $\sigma_{\mathrm{tr}}$ in the Stage~II refinement cycle on the MF2K evaluation set.
Fig.~\ref{fig:sigma_figure} reveals a clear trade-off.
A small $\sigma_{\mathrm{tr}}$ tightly constrains the refinement, so the model edits conservatively, preserving identity but recovering few fine-grained textures and thus lowering MSim.
A larger $\sigma_{\mathrm{tr}}$ relaxes this constraint and raises MSim, but the weaker structural guidance lowers both identity similarity and background stability (higher L2-M).
We therefore set $\sigma_{\mathrm{tr}}=0.6$ by default, which provides a strong balance between cosmetic fidelity and identity preservation while achieving the best background stability.

\begin{figure}[!t]
    \centering
    \includegraphics[width=\linewidth]{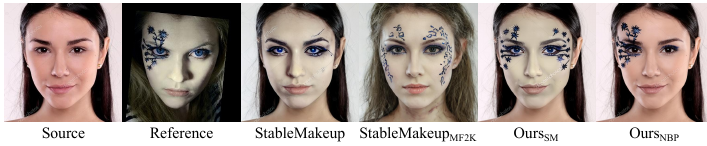}
    \caption{\textbf{Effect of the pseudo-target source and training data.}
        Even from weaker pseudo-targets, Ours$_{\mathrm{SM}}$ matches Ours$_{\mathrm{NBP}}$ in recovering the fine-grained makeup, whereas StableMakeup fails even when retrained on MF2K (StableMakeup$_{\mathrm{MF2K}}$), showing that the reality-anchored refinement breaks the pseudo-target ceiling.}
    \label{fig:ablation_pseudo_target_source}
\end{figure}

\noindent{\textbf{Impact of pseudo-target supervision and data.}}
Tab.~\ref{tab:ablation_teacher} and Fig.~\ref{fig:ablation_pseudo_target_source} analyze how the pseudo-target source and the training data affect transfer.
We train our framework on pseudo-targets from either Nano~Banana~Pro (Ours$_{\mathrm{NBP}}$, our default) or StableMakeup (Ours$_{\mathrm{SM}}$).
Although StableMakeup produces lower-quality pseudo-targets, Ours$_{\mathrm{SM}}$ still improves markedly over StableMakeup itself (8.56 vs.\ 6.73 in MF2K MSim$_G$) and nearly matches Ours$_{\mathrm{NBP}}$, showing that the refinement re-anchors the model to real references rather than being capped by its initial pseudo-targets.
We then retrain StableMakeup on our MF2K data (StableMakeup$_{\mathrm{MF2K}}$) to isolate the contribution of the dataset itself.
Its improvements are limited to simpler styles (MT, MT-Wild) and reverse on the complex makeup of LADN and MF2K, where StableMakeup's global CLIP conditioning cannot represent such fine-grained detail.
Overall, these results confirm that, while higher-quality pseudo-targets help, the reality-anchored refinement is the key to surpassing them and breaking the pseudo-target ceiling.

\begin{table*}[t]
    \centering
    \caption{\textbf{Ablation of the pseudo-target source and training data.}
        StableMakeup$_{\mathrm{MF2K}}$ denotes StableMakeup trained on our MF2K data.
        Ours$_{\mathrm{SM}}$ and Ours$_{\mathrm{NBP}}$ denote our models trained with pseudo-targets synthesized by StableMakeup and Nano~Banana~Pro, respectively.
        }
    \scriptsize
    \setlength{\tabcolsep}{1.4pt}
    \renewcommand{\arraystretch}{1.2}
    \begin{tabular}{l|ccccc|ccccc}
        \toprule
        & \multicolumn{5}{c|}{\textbf{MT Test Set~\cite{li2018beautygan}}} & \multicolumn{5}{c}{\textbf{LADN Test Set~\cite{Gu_2019_ICCV}}} \\
        \cmidrule(lr){2-6}\cmidrule(lr){7-11}
        \multirow{-2.6}{*}{\makecell[c]{\textbf{Methods}}}
        & MSim$_G$$\uparrow$ & MSim$_Q$$\uparrow$ & ID$\uparrow$ & L2-M$\downarrow$ & FID$\downarrow$
        & MSim$_G$$\uparrow$ & MSim$_Q$$\uparrow$ & ID$\uparrow$ & L2-M$\downarrow$ & FID$\downarrow$ \\
        \midrule
        StableMakeup~\cite{zhang2025stablemakeup}
        & 8.34 & 7.30 & .49 & 9.55 & \textbf{82.70}
        & 7.81 & 7.89 & .47 & 8.91 & \textbf{65.99} \\
        StableMakeup$_{\mathrm{MF2K}}$
        & 8.47 & \underline{7.47} & .60 & 30.59 & 88.22
        & 7.47 & 7.75 & .52 & 51.60 & 74.67 \\
        Ours$_{\mathrm{SM}}$
        & \underline{8.83} & \textbf{7.53} & \underline{.65} & \underline{4.79} & \underline{87.07}
        & \underline{9.07} & \textbf{8.46} & \underline{.61} & \underline{4.24} & \underline{67.68} \\
        Ours$_{\mathrm{NBP}}$
        & \textbf{8.96} & 7.12 & \textbf{.87} & \textbf{3.94} & 87.72
        & \textbf{9.14} & \underline{8.04} & \textbf{.81} & \textbf{3.54} & 68.67 \\
        \midrule
        & \multicolumn{5}{c|}{\textbf{MT-Wild Test Set~\cite{jiang2020psgan}}} & \multicolumn{5}{c}{\textbf{Our MF2K Test Set}} \\
        \cmidrule(lr){2-6}\cmidrule(lr){7-11}
        \multirow{-2.6}{*}{\makecell[c]{\textbf{Methods}}}
        & MSim$_G$$\uparrow$ & MSim$_Q$$\uparrow$ & ID$\uparrow$ & L2-M$\downarrow$ & FID$\downarrow$
        & MSim$_G$$\uparrow$ & MSim$_Q$$\uparrow$ & ID$\uparrow$ & L2-M$\downarrow$ & FID$\downarrow$ \\
        \midrule
        StableMakeup~\cite{zhang2025stablemakeup}
        & 8.21 & \textbf{7.62} & .48 & 9.91 & \textbf{86.51}
        & 6.73 & 7.04 & .43 & 9.26 & \underline{100.60} \\
        StableMakeup$_{\mathrm{MF2K}}$
        & 8.66 & \underline{7.62} & .42 & 21.48 & 90.12
        & 6.58 & 6.92 & \underline{.51} & 28.17 & 100.67 \\
        Ours$_{\mathrm{SM}}$
        & \textbf{8.99} & 7.45 & \underline{.75} & \underline{5.46} & \underline{88.60}
        & \underline{8.56} & \textbf{8.16} & \underline{.51} & \underline{5.90} & \textbf{95.84} \\
        Ours$_{\mathrm{NBP}}$
        & \underline{8.77} & 7.00 & \textbf{.85} & \textbf{4.37} & 91.11
        & \textbf{9.22} & \underline{7.68} & \textbf{.74} & \textbf{4.31} & 103.27 \\
        \bottomrule
    \end{tabular}
    \label{tab:ablation_teacher}
\end{table*}

\subsection{{Resolution Scaling}}
\begin{figure}[!t]
    \centering
    \includegraphics[width=\linewidth]{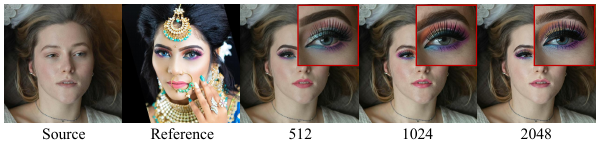}
    \caption{\textbf{Resolution scaling examples.}
        We show 512/1024/2048 outputs with aligned zoom-in crops.
        Higher resolution recovers finer and sharper makeup details.}
    \label{fig:scaling}
\end{figure}

To explore high-resolution portrait makeup transfer, we perform experiments at spatial resolutions of 512, 1024, and 2048 on our MF2K dataset.
To supervise fine details in high-resolution training, we add a wavelet-based loss~\cite{zhang2025diffusion} that uses a discrete wavelet transform (DWT) to emphasize high-frequency detail while preserving low-frequency structure.
Formally, it is defined as $\mathcal{L}_{\mathrm{WLF}} = \mathbb{E} \left[ w_t \| f(v_{\theta}(z_t, t)) - f(\epsilon - x_0) \|^2 \right]$, where $w_t$ is the loss weight, and $f(\cdot)$ denotes the DWT.
As shown in Fig.~\ref{fig:scaling}, increasing the resolution progressively recovers finer and sharper makeup details.
To our knowledge, this is the first high-fidelity makeup transfer at 2K resolution, directly meeting the high-resolution demands of real-world applications.

\subsection{Limitations and Failure Cases}
\label{sec:limitations}

Although our framework faithfully transfers visual appearance cues, it does not explicitly enforce physical lighting and reflectance consistency.
Consequently, significant illumination mismatches can yield contradictory results, such as retaining intense specular highlights from a directionally lit reference within a softly lit source environment.
A detailed discussion of limitations is provided in Suppl.~Sec.~\ref{sec:suppl_limitations}.

\section{Conclusion}
\label{sec:conclusion}

In this paper, we presented ART, a two-stage framework for high-fidelity makeup transfer that breaks the pseudo-target ceiling.
Through a reality-anchored refinement cycle, the transfer model recovers fine-grained makeup textures and precise placement by using the reconstruction of the real reference as its supervisory signal, while a controlled-noise bottleneck stabilizes the refinement and preserves global structure.
We further introduced MakeupFaces2K, a 2K-resolution in-the-wild makeup portrait dataset that supports research on high-resolution makeup transfer and related facial editing tasks.
Extensive experiments show that our method delivers the best makeup fidelity and background stability while robustly preserving identity.
Beyond makeup transfer, the proposed reality-anchored refinement offers a general paradigm for DiT-based image editing.

\section*{Acknowledgements}
This work was supported by the National Key Research and Development Program of China under Grant No. 2022YFA1004100.

\bibliographystyle{splncs04}
\bibliography{main}

\clearpage
\input{suppl}

\end{document}

%% file: suppl.tex
\begin{center}
{\Large\bfseries Anchoring on Reality: Breaking the Pseudo-Target Ceiling in Makeup Transfer\par}
{\large\bfseries Supplementary Material\par}
\end{center}

\renewcommand{\thesection}{\Alph{section}}
\renewcommand{\theHsection}{supp.\Alph{section}}
\renewcommand{\thefigure}{\Alph{figure}}
\renewcommand{\thetable}{\Alph{table}}

\setcounter{section}{0}
\setcounter{figure}{0}
\setcounter{table}{0}

\noindent This supplementary material provides comprehensive implementation details, visual ablations, and extensive qualitative results to further support the claims made in the main paper.
Specifically, it contains:

\begin{itemize}
    \item \textbf{Section~\ref{sec:visual_showcase}: More Qualitative Results.} This section presents additional qualitative results, including high-resolution 2K synthesis, direct evidence of breaking the pseudo-target ceiling, cross-domain generalization (e.g., anime and 3D avatars), and an analysis of the auxiliary makeup remover.
    \item \textbf{Section~\ref{sec:method_ablations}: More Details of the Methodology.} This section presents visual ablations of the controlled-noise bottleneck $\sigma_{\mathrm{tr}}$ and the structural regularizer $\mathcal{L}_{\mathrm{bottleneck}}$, additional implementation details, computational cost, and the effectiveness of unpaired data during refinement.
    \item \textbf{Section~\ref{sec:dataset_eval}: Dataset and Evaluation Details.} This section details the MF2K dataset curation pipeline, discusses limitations of global feature-based metrics, outlines our multi-VLM evaluation and user study protocols.
    \item \textbf{Section~\ref{sec:suppl_limitations}: Limitations and failure cases.} This section presents failure cases and limitations of our framework.
\end{itemize}

\section{Additional Qualitative Results}
\label{sec:visual_showcase}
In this section, we provide extensive qualitative results to further demonstrate the generative capabilities and fine-grained detail preservation of the proposed Anchoring on Reality Makeup Transfer (ART) framework.

\subsection{High-Resolution 2K Synthesis}
\label{subsec:2k_results}

\begin{figure}[!t]
    \centering
    \includegraphics[width=.925\linewidth]{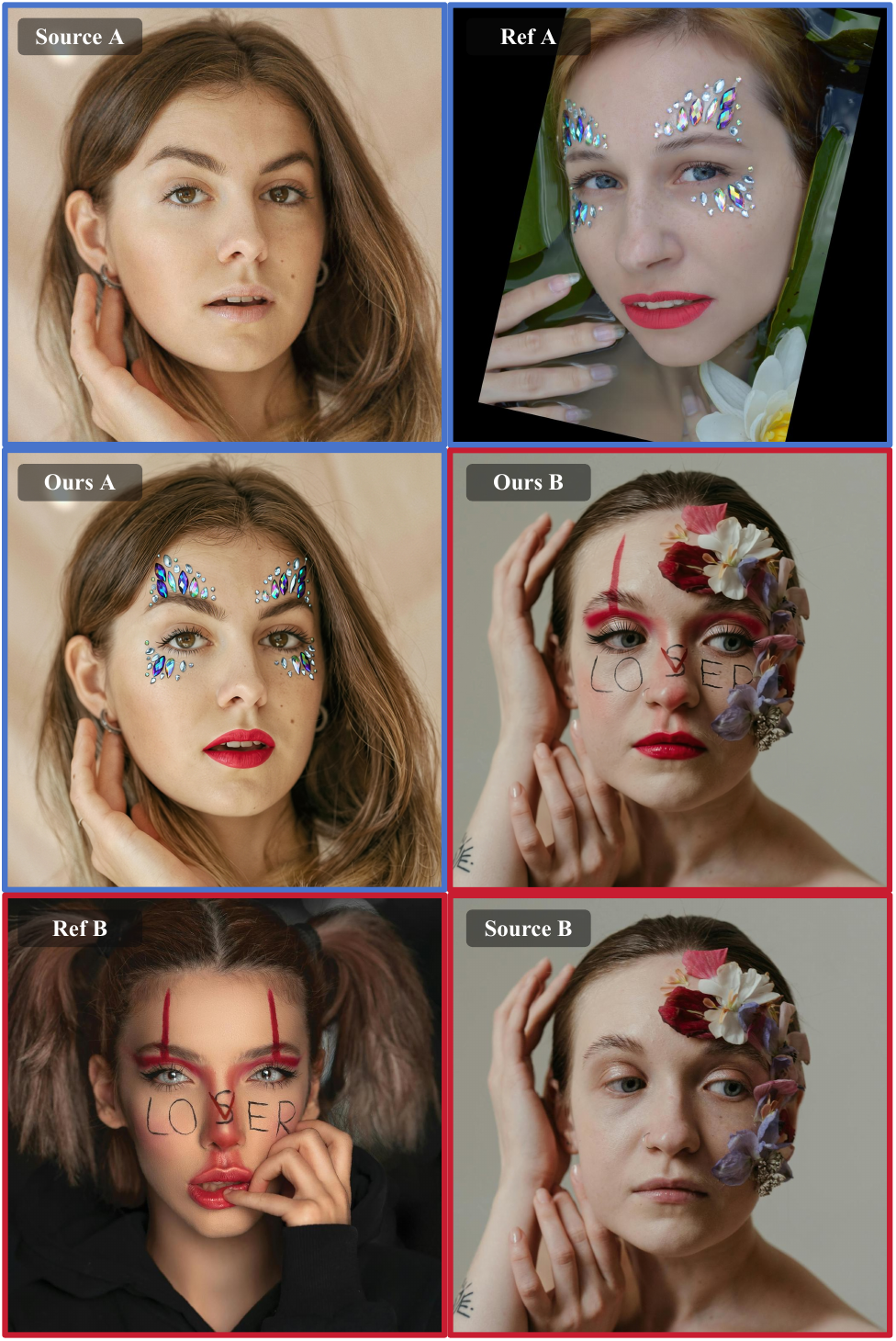}
    \caption{2K-resolution ($2048 \times 2048$) makeup transfer results on our MF2K dataset.
    The \textcolor{blue}{blue}-framed triplet shows example A, and the \textcolor{red}{red}-framed triplet shows example~B.
    All source, reference, and our output images are presented at 2K resolution.}
    \label{fig:supp_2k}
\end{figure}

The MakeupFaces2K (MF2K) dataset is curated to support high-fidelity cosmetic synthesis with rich fine-grained details.
With this dataset, our ART framework scales to 2K resolution ($2048 \times 2048$).
Fig.~\ref{fig:supp_2k} shows high-resolution transfer results.
Our model faithfully preserves challenging cosmetic textures such as individual eyelashes, glitter, and complex face-painting boundaries.

\subsection{More Visual Results}
\label{subsec:more_visual_results}
\begin{figure}[!t]
    \centering
    \includegraphics[width=\linewidth]{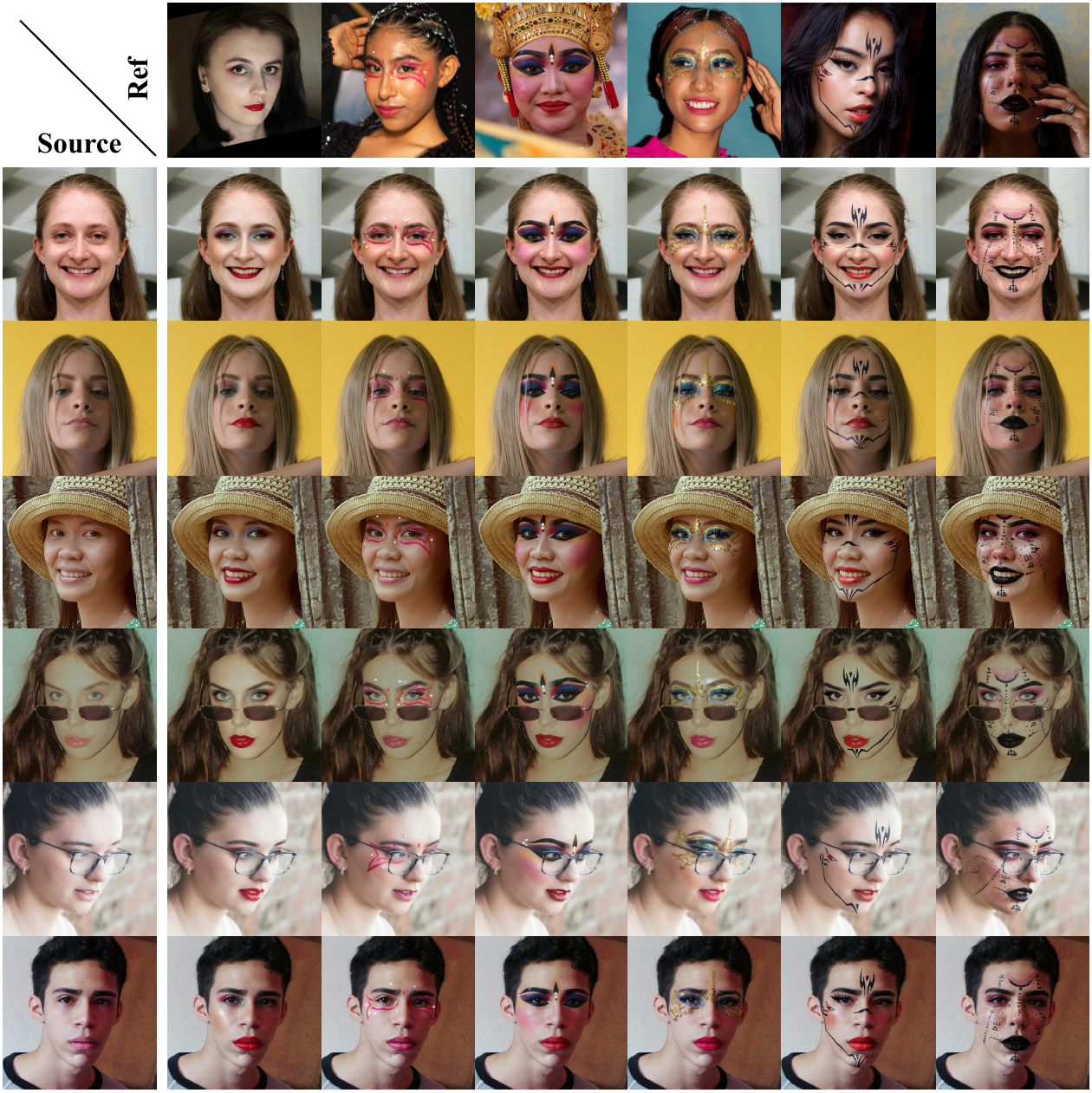}
    \caption{Additional visual results of our ART on diverse makeup styles and facial appearances.
    ART produces photorealistic transfers while preserving the source identity and facial structure, demonstrating the effectiveness of our method.}
    \label{fig:supp_more_visual_results}
\end{figure}

Fig.~\ref{fig:supp_more_visual_results} provides additional visual results of our ART across diverse makeup styles and facial appearances, demonstrating its effectiveness in producing photorealistic transfers while preserving the source identity and facial structure.

\subsection{Breaking the Pseudo-Target Ceiling}
\label{subsec:ceiling_break}
\begin{figure*}[!t]
    \centering
    \includegraphics[width=\linewidth]{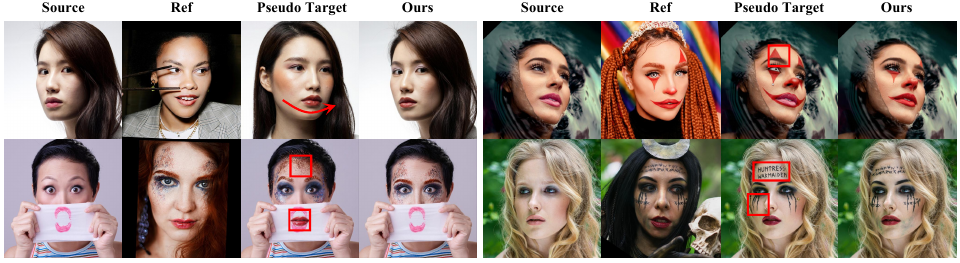}
    \caption{We compare the synthetic pseudo-targets generated by Nano Banana Pro (used for Stage I initialization) against our final ART outputs.
    As highlighted by the red markers, the pseudo-targets frequently suffer from generative flaws, including unintended head pose alterations (top-left), wrong patterns (top-right), and erroneous semantic hallucinations (bottom).
    In contrast, our ART results faithfully recover accurate cosmetic details while strictly preserving the source geometry.}
    \label{fig:supp_ceiling}
\end{figure*}

To validate that ART breaks the ``pseudo-target ceiling'' rather than merely memorizing training data, we conduct inference directly on the training set.
Fig.~\ref{fig:supp_ceiling} compares our final outputs against the synthetic pseudo-targets (generated by Nano Banana Pro~\cite{deepmind_gemini_image_pro}) used for Stage I initialization.
As highlighted by the red markers, these pseudo-targets contain inherent generative flaws, such as altered head poses, incorrect makeup textures, and erroneous facial attributes.
Even when inferencing on these exact training pairs, our ART successfully reconstructs sharper cosmetic textures and corrects the structural artifacts present in the pseudo-targets.
This demonstrates that the differentiable reality-anchored cycle actively overrides synthetic artifacts by extracting direct supervision from the real reference images.

\subsection{Cross-Domain Transfer: Beyond Human Portraits}
\label{subsec:cross_domain}
\begin{figure}[!t]
    \centering
    \includegraphics[width=\linewidth]{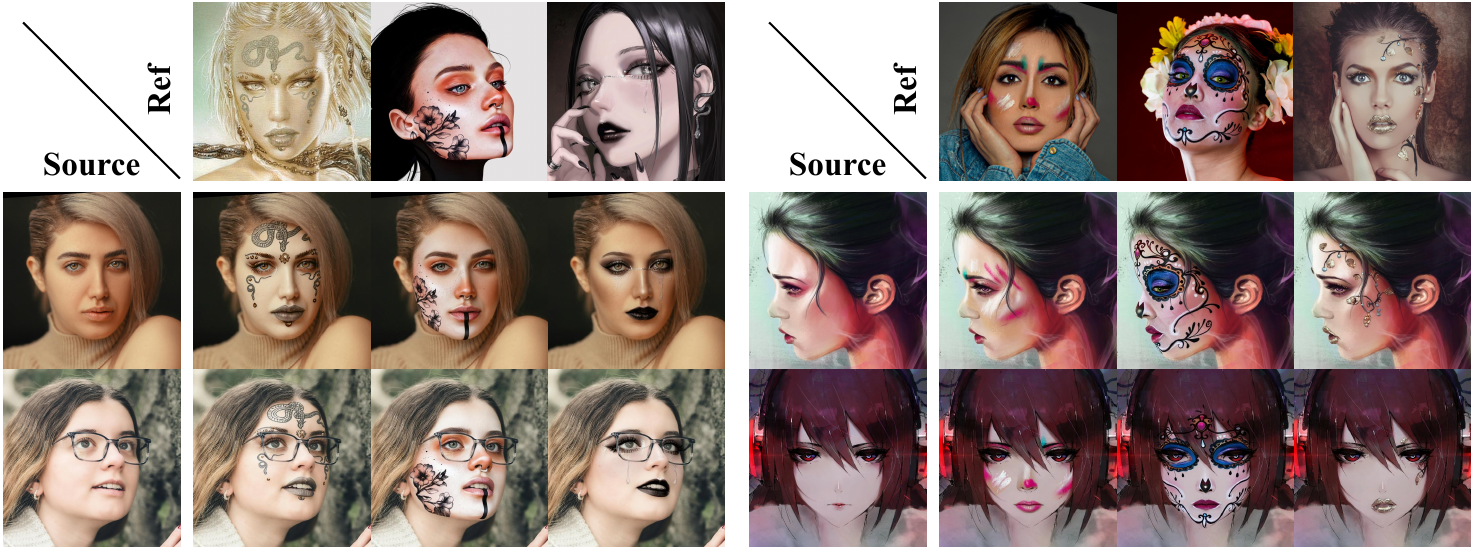}
    \caption{Cross-domain makeup transfer results.
    Despite being trained exclusively on real human portraits, our ART framework exhibits strong zero-shot generalization capabilities on out-of-domain inputs.
    The model can map complex cosmetic semantics across distinct facial topologies without corrupting the original structural geometry.}
    \label{fig:supp_cross_domain}
\end{figure}

To explore the generalization bounds of our ART framework, we evaluate its performance on cross-domain scenarios, including stylized anime characters and 3D avatars.
Although trained exclusively on real human portraits, the semantic alignment established in Stage I, coupled with the reality-anchored refinement cycle in Stage II, endows the model with strong out-of-domain generalizability.
Fig.~\ref{fig:supp_cross_domain} illustrates that ART can effectively map complex cosmetic patterns onto distinct facial topologies without corrupting the original structural geometry.

\subsection{Auxiliary Makeup Remover as a Structural Anchor}
\label{subsec:makeup_removal}
\begin{figure*}[!t]
    \centering
    \includegraphics[width=\linewidth]{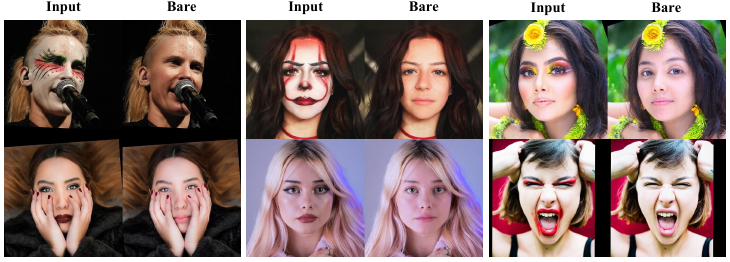}
    \caption{Intermediate outputs of our auxiliary one-step makeup remover $\mathcal{R}$.
    Occasional slight blurriness or minor residues act as an implicit regularization constraint: they force the refinement cycle to extract only the missing high-frequency cosmetic textures from the continuous makeup carrier ($\hat{z}$) and strictly no identity features, thereby preventing shortcut learning and enforcing the disentanglement of identity and makeup.}
    \label{fig:supp_makeup_removal}
\end{figure*}

Our two-stage framework relies on a one-step auxiliary makeup remover $\mathcal{R}$ to extract a bare-skin counterpart ($I_{\mathrm{bare\_ref}}$) from the reference image, as visualized in Fig.~\ref{fig:supp_makeup_removal}.
Optimized primarily for structural and identity preservation, $\mathcal{R}$ isolates the underlying facial geometry.

Because $\mathcal{R}$ operates as an efficient one-step model, its outputs occasionally exhibit slight blurriness or minor cosmetic residues.
Rather than hindering Stage II refinement, these imperfections act as an implicit regularization feature that mechanistically prevents shortcut learning.
During the reality-anchored reconstruction, if the network attempts to compensate for the blurriness of $I_{\mathrm{bare\_ref}}$ by extracting structural details from the continuous makeup carrier $\hat{z}$, it will erroneously reconstruct the source identity, thereby triggering a severe penalty from the refinement loss $\mathcal{L}_{\mathrm{refine}}$ (defined in Eq. 5).
Furthermore, minor cosmetic residues naturally prompt $\hat{z}$ to exclusively encode the remaining high-frequency cosmetic textures required for the full reconstruction.
Consequently, the network is strictly constrained to use $I_{\mathrm{bare\_ref}}$ solely as an identity anchor, guaranteeing that $\hat{z}$ functions purely as a makeup carrier without entangling identity features.

\noindent \textbf{Robustness to Imperfect Removal.}
To quantitatively verify that ART tolerates an imperfect bare-skin anchor, we simulate residual cosmetics by blending the remover output with the reference, $\tilde{I}_{\mathrm{bare\_ref}}=(1-\alpha)I_{\mathrm{bare\_ref}}+\alpha I_{\mathrm{ref}}$, where a larger $\alpha$ leaves stronger makeup residue.
As reported in Tab.~\ref{tab:supp_remover_sensitivity} (averaged over the four test sets), even with a notable residue of $\alpha=0.10$, all metrics remain competitive and degrade only marginally, demonstrating that Stage II refinement is robust to imperfect removal and degrades gracefully even under stronger-than-natural residue.

\begin{table}[!t]
    \centering
    \caption{Sensitivity of ART to imperfect makeup removal, simulated by injecting residual makeup with ratio $\alpha$ into the bare-skin reference.
    Results are averaged over the four test sets.}
    \label{tab:supp_remover_sensitivity}
    \scriptsize
    \setlength{\tabcolsep}{10pt}
    \renewcommand{\arraystretch}{1.1}
    \begin{tabular}{c|cccc}
    \toprule
    $\alpha$ & MSim$_G$$\uparrow$ & MSim$_Q$$\uparrow$ & ID$\uparrow$ & L2-M$\downarrow$ \\
    \midrule
    0.00 & \textbf{9.02} & \textbf{7.46} & \textbf{0.82} & \textbf{4.04} \\
    0.05 & \underline{8.90} & \underline{7.41} & \textbf{0.82} & \underline{4.65} \\
    0.10 & 8.84 & 7.38 & 0.80 & 4.67 \\
    \bottomrule
    \end{tabular}
\end{table}

\section{Method Details and Additional Ablations}
\label{sec:method_ablations}
This section delves into the visual impact of key hyper-parameters and architectural designs, offering a deeper understanding of the inner workings of our reality-anchored refinement.

\subsection{Visual Ablation of the Controlled-Noise Bottleneck ($\sigma_{\mathrm{tr}}$)}
\label{subsec:ablation_sigma}
To evaluate the impact of the controlled noise level $\sigma_{\mathrm{tr}}$ (defined in Eq. 4), we provide a comprehensive visual analysis by directly decoding the continuous makeup carrier $\hat{z}$ (defined in Eq. 4) into the pixel space.

\begin{figure*}[!t]
    \centering
    \includegraphics[width=\linewidth]{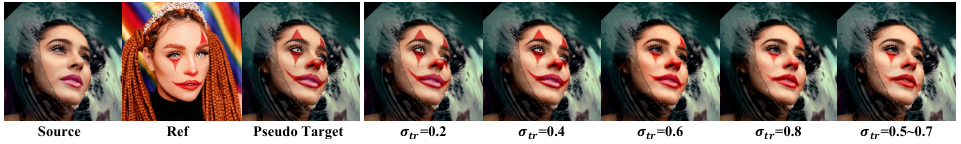}
    \caption{Visual ablation of varying the controlled noise level ($\sigma_{\mathrm{tr}}$) at 20K steps.
    At lower values ($\sigma_{\mathrm{tr}}=0.2$), the model fails to escape the synthetic prior, retaining generative hallucinations from the pseudo-target, such as the extra red diamond on the forehead.
    At $\sigma_{\mathrm{tr}}=0.6$, it balances identity preservation with high-fidelity cosmetic recovery, accurately restoring the sharply bounded, asymmetrical clown eye paint.
    The dynamic range ($\sigma_{\mathrm{tr}} \in [0.5, 0.7]$) shows comparable effectiveness.}
    \label{fig:supp_sigma_val}
\end{figure*}

\noindent \textbf{Impact of Controlled Noise Levels.} Fig.~\ref{fig:supp_sigma_val} evaluates the decoded makeup carrier at 20K training steps across varying $\sigma_{\mathrm{tr}}$ values.
When $\sigma_{\mathrm{tr}}$ is small (e.g., $0.2$), the carrier is overly constrained by the synthetic prior, visibly retaining generative hallucinations from the pseudo-target, such as the extra red diamond makeup on the forehead and the accidentally altered pupil color.
Conversely, a higher value (e.g., $0.8$) grants excessive generative freedom, introducing identity and background drifts and blurring high-frequency details.
Setting $\sigma_{\mathrm{tr}}=0.6$ provides an optimal generative balance: it effectively overrides the pseudo-target's artifacts while accurately extracting the sharply bounded, asymmetrical clown eye paint directly from the real reference.
We also visualized training with a dynamic sampling range ($\sigma_{\mathrm{tr}} \in [0.5, 0.7]$), which yields comparable visual results.
For simplicity and to maintain a consistent balance between identity preservation and makeup fidelity, we adopt a fixed $\sigma_{\mathrm{tr}}=0.6$ in our standard pipeline.

\begin{figure*}[!t]
    \centering
    \includegraphics[width=\linewidth]{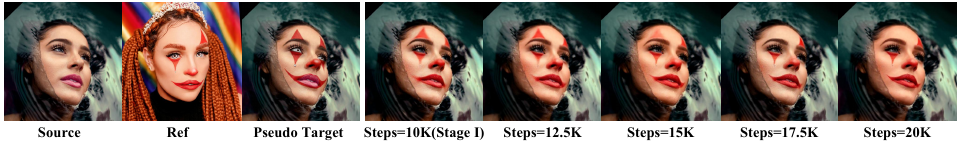}
    \caption{Temporal evolution of the makeup carrier ($\hat{z}$) at fixed $\sigma_{\mathrm{tr}}=0.6$.
    Starting from the pseudo-target's initialization at 10K steps (which contains blurred artifacts and mismatched geometries), the reality-anchored refinement progressively corrects the erroneous makeup into detailed and well-aligned patterns (20K steps), demonstrating the active overriding of synthetic artifacts during training.}
    \label{fig:supp_sigma_steps}
\end{figure*}

\noindent \textbf{Training Dynamics of $\hat{z}$.} Fig.~\ref{fig:supp_sigma_steps} illustrates the temporal evolution of the makeup carrier at $\sigma_{\mathrm{tr}}=0.6$.
At 10K steps (the end of Stage I), the carrier inherits the blurred artifacts and mismatched geometries from the pseudo-target.
As Stage II progresses (12.5K to 20K steps), the $\mathcal{L}_{\mathrm{refine}}$ objective (Eq. 5) actively drives $\hat{z}$ to extract accurate cosmetic cues from the reference image.
The erroneous makeup is progressively corrected into detailed and well-aligned patterns, providing empirical evidence that the framework breaks the pseudo-target ceiling over training.

\subsection{Ablation of Structural Regularization}
\label{subsec:ablation_bottleneck_loss}
\begin{figure*}[!t]
    \centering
    \includegraphics[width=\linewidth]{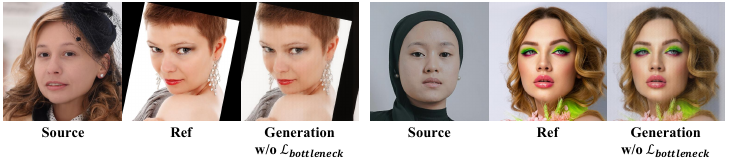}
    \caption{Analyses without the structural regularizer $\mathcal{L}_{\mathrm{bottleneck}}$.
    When optimized solely with $\mathcal{L}_{\mathrm{refine}}$, the network takes a shortcut by ignoring the source conditioning and directly outputting the reference image to trivially minimize the reconstruction loss.}
    \label{fig:supp_degenerate}
\end{figure*}

In the main paper, we emphasized the necessity of the structural regularizer $\mathcal{L}_{\mathrm{bottleneck}}$ (Eq. 6) to ensure the model preserves the source facial structure during the refinement process.
Fig.~\ref{fig:supp_degenerate} visually demonstrates the failure case that occurs when this regularization is removed ($\lambda_{\mathrm{bot}}=0$).
Without $\mathcal{L}_{\mathrm{bottleneck}}$ anchoring the initial transfer prediction to the source geometry, the makeup carrier $\hat{z}$ becomes unconstrained.
To minimize the reconstruction loss $\mathcal{L}_{\mathrm{refine}}$ (Eq. 5), the network takes a ``shortcut'' by forcing $\hat{z}$ to encode the entire reference image rather than disentangling its cosmetic semantics.
Consequently, it entirely discards the source identity and simply outputs a copy of the reference.
Incorporating $\mathcal{L}_{\mathrm{bottleneck}}$ effectively restricts this behavior, compelling the model to extract only cosmetic textures while maintaining the original structure.

\begin{figure*}[!t]
    \centering
    \includegraphics[width=\linewidth]{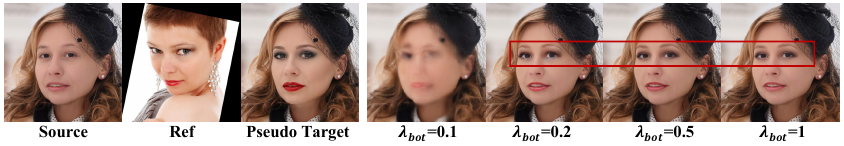}
    \caption{Visual ablation of the structural regularizer weight ($\lambda_{\mathrm{bot}}$).
    A weak penalty ($\lambda_{\mathrm{bot}}=0.1$) causes severe blurring.
    An excessive penalty ($\lambda_{\mathrm{bot}} \ge 0.5$) overly binds the network to the pseudo-target prior, failing to recover the vivid, high-frequency eye makeup present in the real reference (as highlighted by the red boxes).}
    \label{fig:supp_ablation_lambda_bot}
\end{figure*}

Furthermore, to justify our empirical choice of the penalty weight $\lambda_{\mathrm{bot}}=0.2$ (defined in Eq. 7), we conduct a comprehensive visual ablation across varying values, as illustrated in Fig.~\ref{fig:supp_ablation_lambda_bot}.
A weak penalty (e.g., $\lambda_{\mathrm{bot}}=0.1$) provides insufficient structural anchoring, resulting in severe blurring and a collapse of facial details.
Conversely, an excessive penalty (e.g., $\lambda_{\mathrm{bot}} \ge 0.5$) overly constrains the generation to the synthetic pseudo-target prior.
As highlighted by the red boxes, these heavy constraints hinder the model's ability to absorb high-frequency textural corrections from the reference, forcing the output to retain the suboptimal cosmetic patterns of the pseudo-target.
Setting $\lambda_{\mathrm{bot}}=0.2$ achieves the optimal generative balance, extracting high-fidelity cosmetic textures from the reality-anchored cycle while preserving the pixel-aligned source geometry.

\subsection{Additional Implementation Details}
\label{subsec:implementation_details}
To optimize memory efficiency, all training is conducted using \texttt{bfloat16} mixed precision coupled with gradient checkpointing.

\begin{figure}[!t]
    \centering
    \includegraphics[width=\linewidth]{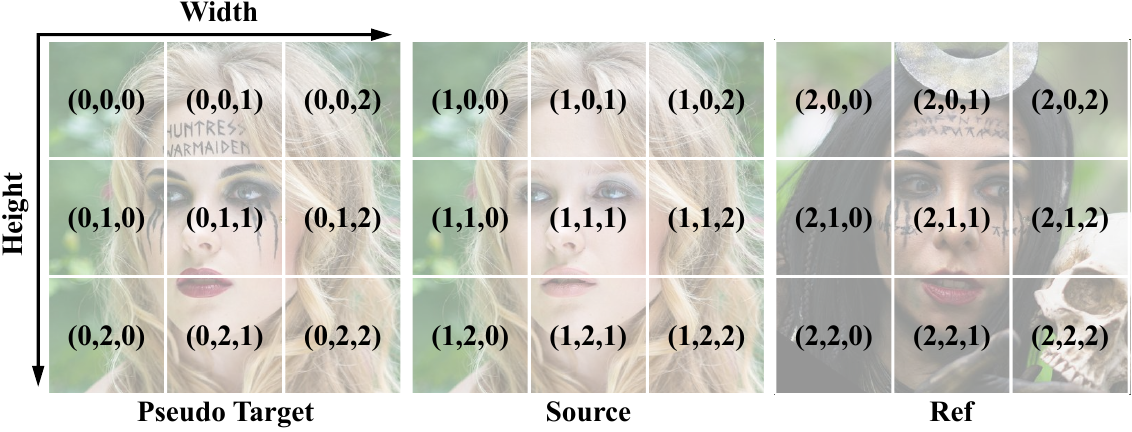}
    \caption{Schematic of the extended 3D Rotary Position Embedding (RoPE) for multi-image condition injection.
    By introducing an explicit image index $i$, tokens from the pseudo-target ($i=0$), source ($i=1$), and reference ($i=2$) are assigned distinct 3D spatial coordinates $(i, y, x)$.
    This prevents spatial collisions within the concatenated sequence and facilitates accurate cross-image semantic attention.}
    \label{fig:supp_rope}
\end{figure}

\noindent \textbf{Multi-Image Condition Injection.} To simultaneously inject the noisy target, source, and reference images into our FLUX.1-Kontext-dev~\cite{batifol2025flux} backbone, we extend its native multi-dimensional Rotary Position Embedding (RoPE).
As illustrated in Fig.~\ref{fig:supp_rope}, each token's spatial coordinate is formulated as $\mathbf{p} = (i, y, x)$, where $i$, $y$, and $x$ denote the image, height, and width axes, respectively.
Within the concatenated latent sequence, we assign the image index $i=0$ to the target, $i=1$ to the source, and $i=2$ to the reference.
This strictly avoids spatial collisions, enabling the joint self-attention mechanism to align intra-image features while distinguishing inter-image semantics.

\begin{figure}[!t]
    \centering
    \includegraphics[width=\linewidth]{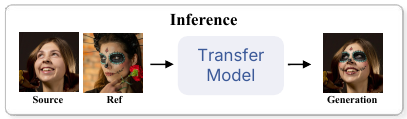}
    \caption{Overview of the ART inference pipeline.
    Unlike the complex reality-anchored cycle used during training, testing operates as a highly efficient, one-pass generation process conditioned solely on the source and reference images.
    No auxiliary makeup remover or pseudo-target synthesis is required.}
    \label{fig:supp_infer_pipeline}
\end{figure}

\noindent \textbf{Inference Pipeline.}
It is crucial to emphasize that the reality-anchored refinement cycle and the auxiliary makeup remover are utilized exclusively during the training phase.
During inference, our ART framework operates as a highly efficient, one-pass generation process (see Fig.~\ref{fig:supp_infer_pipeline}).
Given a source image $I_{\mathrm{src}}$ and a reference image $I_{\mathrm{ref}}$, we extract their latent representations ($z_{\mathrm{src}}$ and $z_{\mathrm{ref}}$) and concatenate them with pure Gaussian noise $z_1 \sim \mathcal{N}(0, I)$.
The final transferred output is then generated by solving the Flow Matching ODE trajectory (defined in Eq. 1) \cite{lipman2022flow} in a single forward pass (using 28 Euler steps) conditioned on $z_{\mathrm{src}}$ and $z_{\mathrm{ref}}$.
No pseudo-target synthesis, iterative optimization, or makeup removal is required during testing, ensuring significant inference efficiency.

\noindent \textbf{Auxiliary Makeup Remover Training.}
The auxiliary makeup remover $\mathcal{R}$ is optimized independently using a subset of the constructed training triplets.
Specifically, it is trained on $(I_{\mathrm{pseudo}}, I_{\mathrm{src}})$ pairs, where $I_{\mathrm{src}}$ strictly comprises bare-skin portraits.
It shares the same backbone architecture as the primary transfer model.
We employ Low-Rank Adaptation (LoRA)~\cite{hu2022lora} for fine-tuning, setting both the rank and the scaling factor to 32.
To facilitate efficient inference and strictly preserve the underlying facial structure, the model is trained under a one-step generation paradigm by fixing the timestep at $t=999$ (pure noise) to directly predict the bare-skin counterpart.
Optimization is performed using the Prodigy optimizer~\cite{mishchenko2023prodigy} with a learning rate of 1.
The training spans 10K steps with a batch size of 4 on a single NVIDIA H20 GPU.
The loss weights defined in Eq. (3) are set to $\lambda_{\mathrm{lpips}}=1$, $\lambda_{\mathrm{id}}=2$, and $\lambda_{\mathrm{lmk}}=10^4$.

\noindent \textbf{Computational Cost.} Tab.~\ref{tab:supp_cost} reports the measured cost of ART.
The additional overhead is confined to training: Stage II is about $1.6\times$ slower per step than Stage I due to the extra reconstruction pass, and the auxiliary remover is trained only once and then frozen.
At inference, ART runs a single transfer pass without the remover, pseudo-target synthesis, or refinement cycle, making it as efficient as a standard one-pass DiT generation.
\begin{table}[!t]
    \centering
    \caption{Computational cost of ART.
    All measurements are on a single NVIDIA H20 GPU with \texttt{bfloat16} precision, LoRA rank 32.}
    \label{tab:supp_cost}
    \scriptsize
    \setlength{\tabcolsep}{10pt}
    \renewcommand{\arraystretch}{1.1}
    \begin{tabular}{ll|cc}
    \toprule
    \textbf{Mode} & \textbf{Component} & \textbf{Time} & \textbf{VRAM} \\
    \midrule
    \multirow{3}{*}{Train} & Stage I  & 8.22\,s/step  & 41.3\,GB \\
    & Stage II & 13.10\,s/step & 46.5\,GB \\
    & Remover  & 5.78\,s/step  & 52.9\,GB \\
    \midrule
    Infer & Transfer only & 11.45\,s/img & 32.4\,GB \\
    \bottomrule
    \end{tabular}
\end{table}

\subsection{Effectiveness of Unpaired Data in Refinement}
\label{subsec:unpaired_data_ablation}
While Stage I relies on 2,000 curated pseudo-triplets to establish the initial mapping, our ART framework is capable of leveraging the remaining $\sim$4{,}500 unpaired images in the MF2K dataset during Stage II refinement.
For these un-paired samples, where no high-quality pseudo-target $I_{\mathrm{pseudo}}$ is available, we substitute $z_{\mathrm{pseudo}}$ with the source latent $z_{\mathrm{src}}$ and set the structural regularizer weight $\lambda_{\mathrm{bot}}=0$ in Eq. (7).
This configuration allows the model to utilize the differentiable reality-anchored cycle as a self-supervised objective on a broader distribution of real-world faces.

As reported in Tab.~\ref{tab:ablation_unpaired}, we compare our full model (trained on the complete MF2K dataset) against a variant trained exclusively on the 2,000 paired triplets.
The results indicate that while the 2,000 paired samples already provide a strong baseline, the inclusion of the additional $\sim$4{,}500 unpaired images yields a consistent improvement in makeup fidelity (MSim$_G$ increasing from 9.14 to 9.22).
This demonstrates that the reality-anchored cycle effectively bridges the gap between synthetic supervision and real-world distribution, allowing the model to further refine fine-grained cosmetic textures by ``observing'' a larger variety of real facial identities without additional pseudo-target synthesis.

\begin{table}[!t]
    \centering
    \caption{Ablation of utilizing additional unpaired images during Stage II refinement.}
    \label{tab:ablation_unpaired}
    \scriptsize
    \setlength{\tabcolsep}{10pt}
    \renewcommand{\arraystretch}{1.1}
    \begin{tabular}{l|ccccc}
    \toprule
    \textbf{Refinement Data} & MSim$_G$$\uparrow$ & MSim$_Q$$\uparrow$ & ID$\uparrow$ & L2-M$\downarrow$ & FID$\downarrow$ \\
    \midrule
    Paired Triplets Only & 9.14 & 7.40 & \textbf{0.74} & \textbf{4.00} & 103.67 \\
    \textbf{Full Dataset(Ours)} & \textbf{9.22} & \textbf{7.68} & \textbf{0.74} & 4.31 & \textbf{103.27} \\
    \bottomrule
    \end{tabular}
\end{table}

\section{Dataset, Evaluation, and User Study Details}
\label{sec:dataset_eval}

\subsection{Details of the MakeupFaces2K (MF2K) Dataset}
\label{subsec:mf2k_details}
\noindent \textbf{Dataset Overview and Demographics.} The MakeupFaces2K (MF2K) dataset comprises 8,573 2K-resolution ($2048 \times 2048$) in-the-wild facial portraits.
To ensure comprehensive coverage, the dataset is categorized into four cosmetic intensities: bare skin (3,139), light makeup (2,063), heavy makeup (1,798), and artistic styles (1,573).
Comprehensive dataset-level statistics are detailed in Tab.~\ref{tab:supp_mf2k_overall}.

\begin{table*}[!t]
    \centering
    \scriptsize
    \caption{Overall statistics of the MakeupFaces2K (MF2K) dataset.
    Demographic labels (Gender and Age Group) are coarse appearance-based annotations used strictly for dataset-level statistical analysis and do not represent self-identified attributes.}
    \label{tab:supp_mf2k_overall}
    \setlength{\tabcolsep}{6pt}
    \renewcommand{\arraystretch}{1.15}
    \begin{tabular}{@{}lccccccccc@{}}
    \toprule
    \multirow{2}{*}{\textbf{Metric}}
    & \multicolumn{4}{c}{\textbf{Makeup Category}}
    & \multicolumn{2}{c}{\textbf{Gender}}
    & \multicolumn{3}{c}{\textbf{Age Group}} \\
    \cmidrule(lr){2-5} \cmidrule(lr){6-7} \cmidrule(lr){8-10}
    & Bare & Light & Heavy & Artistic
    & Female & Male
    & Child & Adult & Elderly \\
    \midrule
    \textbf{Count}
    & 3139 & 2063 & 1798 & 1573
    & 6726 & 1847
    & 758 & 7605 & 210 \\
    \textbf{Ratio(\%)}
    & 36.61 & 24.06 & 20.97 & 18.35
    & 78.46 & 21.54
    & 8.84 & 88.71 & 2.45 \\
    \bottomrule
    \end{tabular}
\end{table*}

\begin{table}[!t]
    \centering
    \scriptsize
    \caption{Data distributions of the Source and Reference pools utilized for pseudo-triplet construction.
    To construct 2,000 diverse training pairs, images were randomly sampled and paired between these two independent pools.}
    \label{tab:supp_pseudo_pools}
    \begin{minipage}[t]{0.59\textwidth}
    \centering
    {\scriptsize \textbf{(a) Source Pool}}
    \setlength{\tabcolsep}{3pt}
    \renewcommand{\arraystretch}{1}
    \begin{tabular}[t]{@{}llccc@{}}
    \toprule
    \textbf{Makeup} & \textbf{Gender} & \textbf{Age} & \textbf{Count} & \textbf{Subtotal} \\
    \midrule
    \multirow{4}{*}{\textbf{Bare}}
    & \multirow{2}{*}{Female} & Adult   & 1000 & \multirow{2}{*}{1050 (52.5\%)} \\
    &                         & Elderly & 50   & \\
    \cmidrule{2-5}
    & \multirow{2}{*}{Male}   & Adult   & 300  & \multirow{2}{*}{350 (17.5\%)} \\
    &                         & Elderly & 50   & \\
    \midrule
    \textbf{Light}& Female & Adult & 500 & 500 (25.0\%) \\
    \midrule
    \textbf{Heavy}& Female & Adult & 100 & 100 (5.0\%) \\
    \bottomrule
    \end{tabular}
    \end{minipage}%
    \hspace{0.02\textwidth}%
    \begin{minipage}[t]{0.39\textwidth}
    \centering
    {\scriptsize \textbf{(b) Reference Pool}}
    \setlength{\tabcolsep}{6pt}
    \renewcommand{\arraystretch}{1.65}
    \begin{tabular}[t]{@{}lcr@{}}
    \toprule
    \textbf{Makeup} & \textbf{Count} & \textbf{Ratio} \\
    \midrule
    \textbf{Light}    & 800 & 40.0\% \\
    \textbf{Heavy}    & 700 & 35.0\% \\
    \textbf{Artistic} & 500 & 25.0\% \\
    \midrule
    \textbf{Total}    & \textbf{2000} & \textbf{100\%} \\
    \bottomrule
    \end{tabular}
    \end{minipage}
\end{table}

\noindent \textbf{Pseudo-Triplet Construction.} To construct the 2,000 training pseudo-triplets, we curated two independent data subsets: a Source Pool and a Reference Pool.
As detailed in Tab.~\ref{tab:supp_pseudo_pools}, the Source Pool (2,000 images) was structured to encompass broad demographic diversity—including variations in age (adult and elderly) and a substantial proportion of male subjects (17.5\%)—while predominantly featuring bare-skin portraits (70\%) to provide a clean geometric foundation.
Furthermore, we intentionally included a minor proportion of portraits with existing cosmetics in the Source Pool to mimic real-world applications where source inputs are rarely perfectly bare-skin.
Conversely, the Reference Pool (2,000 images) was curated to provide rich cosmetic guidance, spanning light (40\%), heavy (35\%), and artistic (25\%) makeup styles.
We then performed randomized bipartite sampling between these two independent pools to formulate structurally and stylistically diverse $(I_{\mathrm{src}}, I_{\mathrm{ref}})$ pairs.
Finally, these pairs were fed into the Nano Banana Pro~\cite{deepmind_gemini_image_pro} to synthesize the corresponding pseudo-targets ($I_{\mathrm{pseudo}}$).

\FloatBarrier
\noindent\begin{minipage}{\linewidth}
\noindent The standardized prompt utilized for this synthesis is provided below:

\begin{lstlisting}
Task: Realistic Makeup Transfer.
Goal: Apply the makeup style from the Reference Image (the second image B) onto the Bare Face Image (the first image A).

1. Target Editing (Applying Makeup to Face A)
Action: Analyze the cosmetic style, colors, textures, and application techniques from the Reference Image and realistically recreate them on the Target Face.

- General Skin Base: Apply the reference's foundation style, finish, and coverage. Adapt the shade to seamlessly blend with the target's natural skin undertone, preserving their original complexion.
- Contour & Highlight: Translate the contouring, highlighting, and blush placement from the reference face onto the bone structure of the Target Face to enhance its existing features.
- Eyes: Recreate the complete eye makeup look. This includes eyeshadow palette and blending gradient, eyeliner style (color, thickness, wing shape), and mascara/false lash style from the reference, adapted to the target's eye shape.
- Brows: Shape, fill, and color the target's eyebrows to match the grooming style and density of the reference brows.
- Lips: Apply the specific lip color, texture (e.g., glossy, velvet matte, creamy), and liner precision from the reference to the target's lips.
- (Optional) Decals/Face Art: If the reference has specific face stickers, jewels, or artistic paint, transfer them to corresponding positions on the target face.

2. Strict Preservation Constraints (Protecting Image A)
Crucial: The identity of Face A must be paramount. The makeup must conform to A's features, not morph A into B.

- Identity & Geometry: The facial anatomy, unique bone structure, nose shape, eye shape, and lip shape of the Target Face (Image A) must remain pixel-perfectly identical. Do not warp features to match the reference.
- Skin Texture: While applying makeup, preserve the underlying realistic skin texture of Target A. The makeup should look like it is sitting *on* real skin, not blurring it into plastic.
- Expression & Pose: Keep the exact mouth openness, teeth visibility, gaze direction, and micro-expressions of Image A.
- Environment & Lighting: The lighting direction, shadows, reflections, background, hair, and accessories from Image A must stay unchanged. The new makeup must interact realistically with Image A's existing lighting.
\end{lstlisting}
\end{minipage}
\FloatBarrier

\noindent \textbf{MF2K Test Set Construction.} To rigorously evaluate our model's capability in transferring highly complex and fine-grained cosmetic styles, we constructed a dedicated evaluation split, termed the MF2K Test Set.
Specifically, we randomly sampled 100 $(I_{\mathrm{src}}, I_{\mathrm{ref}})$ pairs from the MF2K dataset.
To pose a challenging evaluation scenario, the source images ($I_{\mathrm{src}}$) were strictly selected from the bare-skin category, while the reference images ($I_{\mathrm{ref}}$) were exclusively drawn from the Artistic makeup category.
Crucially, we enforce a strict separation between the training and testing sets; all 200 unique images involved in this test set are strictly held out with no overlap with the training data or the pseudo-triplet pools, ensuring a zero-shot evaluation setting.

\begin{figure}[!t]
    \centering
    \includegraphics[width=\linewidth]{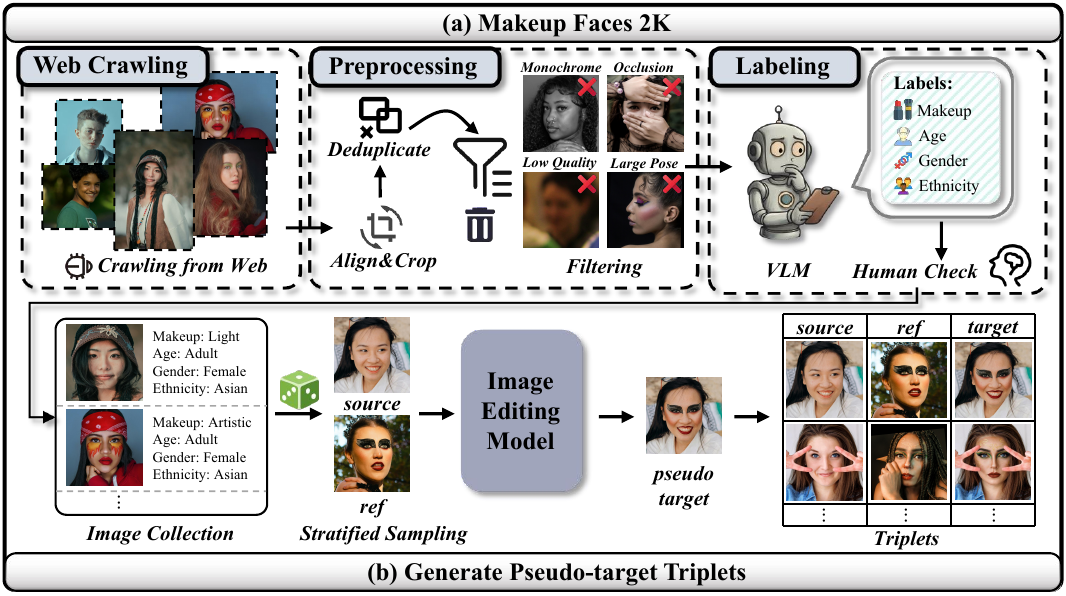}
    \caption{Overview of the MF2K dataset construction and pseudo-triplet generation pipeline. \textbf{Top (Data Curation):} The process strictly filters raw web images through precise alignment, cropping, CLIP-based deduplication, IQA scoring, Euler-angle-based pose filtering and VLM-assisted human-reviewed annotation. \textbf{Bottom (Triplet Construction):} Filtered images are sampled according to predefined stratified ratios to form Source-Reference pairs.
    Finally, Nano Banana Pro is queried to synthesize the pseudo-target ($I_{\mathrm{pseudo}}$), yielding the final training triplets.}
    \label{fig:supp_data_construction}
\end{figure}

\noindent \textbf{Data Curation Pipeline.} Fig.~\ref{fig:supp_data_construction} illustrates our complete data curation and pseudo-triplet construction pipeline.
Initially, we collected approximately 100,000 raw high-resolution portrait images from the internet, supplemented by original high-quality images in Flickr from the FFHQ dataset~\cite{karras2019style}.
To guarantee high-fidelity standards, these images underwent a rigorous preprocessing pipeline.
First, we performed precise facial alignment and cropping.
Second, we applied CLIP-based~\cite{radford2021learning} feature deduplication to eliminate highly similar images early in the pipeline.
Third, to ensure pristine texture quality and spatial stability, we utilized the \texttt{topiq\_nr-face}~\cite{chen2024topiq} metric for Image Quality Assessment (IQA), retaining only images with a score $\ge 0.8$.
Furthermore, we extracted Euler angles using 3DDFA-V2~\cite{guo2020towards}, discarding portraits with a pitch exceeding $\pm 30^\circ$, a yaw exceeding $\pm 45^\circ$, or a roll exceeding $\pm 20^\circ$.
Finally, we employed Qwen2.5-VL-72B-Instruct~\cite{bai2025qwen25vltechnicalreport} to automatically annotate metadata, which was subsequently subjected to human expert review.
Note that demographic labels are coarse, appearance-based annotations utilized strictly for dataset-level statistical balancing and do not represent self-identified demographic attributes.

\subsection{Analysis of Feature-based Metrics and VLM Evaluation}
\label{subsec:metric_failures}
\begin{figure*}[!t]
    \centering
    \includegraphics[width=\linewidth]{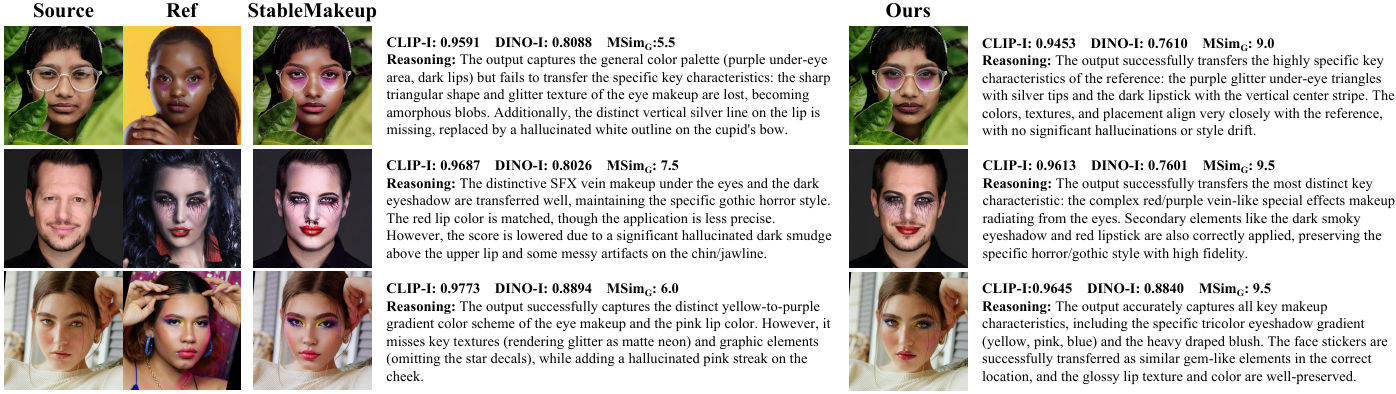}
    \caption{Visual analysis of feature-based metrics.
    While CLIP-I and DINO-I effectively capture global color palettes, they are less sensitive to fine-grained geometric mismatches or occasional artifacts.
    In contrast, the VLM metric (MSim$_G$) explicitly reasons about local textures, key characteristics, and structural preservation, aligning more closely with human perception for complex makeup.}
    \label{fig:supp_metric_failures}
\end{figure*}

Prior makeup transfer works frequently employ feature-based distances (e.g., CLIP-I~\cite{radford2021learning}, DINO-I~\cite{oquab2023dinov2, simeoni2025dinov3}) to quantitatively evaluate makeup similarity.
While these metrics are highly effective at capturing global style and color consistency, we observe that they may exhibit sub-optimal alignment with human perception when evaluating complex, fine-grained cosmetic styles.
Specifically, Fig.~\ref{fig:supp_metric_failures} demonstrates that baseline models can receive high CLIP-I and DINO-I scores by successfully capturing the overall global color palette (e.g., the purple hue).
However, these global embeddings are inherently less sensitive to localized high-frequency semantics, such as the precise shapes of geometric textures (e.g., star decals, glitters) or spatial layouts.
Furthermore, DINO features can occasionally entangle facial structural identity with dense cosmetic patterns, which may lead to biased evaluations when transferring extreme artistic face paintings that alter perceived geometric boundaries.

\begin{table*}[!t]
    \centering
    \caption{Quantitative results of our model and baselines evaluated by feature-based metrics (CLIP-I and DINO-I) alongside our VLM-based evaluation metrics (MSim$_G$ and MSim$_Q$).
    The best and second best results are highlighted in \textbf{bold} and \underline{underline}.}
    \label{tab:supp_feature_metrics}
    \scriptsize
    \setlength{\tabcolsep}{2.5pt}
    \renewcommand{\arraystretch}{1.08}
    \begin{tabular}{l|cccc|cccc}
    \toprule
    & \multicolumn{4}{c|}{\textbf{MT Test Set~\cite{li2018beautygan}}} & \multicolumn{4}{c}{\textbf{LADN Test Set~\cite{Gu_2019_ICCV}}} \\
    \cmidrule(lr){2-5}\cmidrule(lr){6-9}
    \multirow{-2.6}{*}{\makecell[c]{\textbf{Methods}}}  &
    CLIP-I$\uparrow$ &
    DINO-I$\uparrow$ &
    MSim$_G$$\uparrow$ &
    MSim$_Q$$\uparrow$ &
    CLIP-I$\uparrow$ &
    DINO-I$\uparrow$ &
    MSim$_G$$\uparrow$ &
    MSim$_Q$$\uparrow$ \\
    \midrule
    PSGAN~\cite{jiang2020psgan}                 & 0.934 & 0.727 & 6.13 & 5.26 & 0.936 & 0.714 & 3.79 & 3.44 \\
    EleGANt~\cite{yang2022elegant}               & 0.938 & 0.720 & 6.98 & 4.94 & 0.942 & 0.726 & 5.64 & 4.82 \\
    MAD~\cite{ruan2025mad}                   & 0.944 & \underline{0.740} & 6.60 & 5.10 & 0.936 & 0.715 & 3.78 & 3.49 \\
    SHMT~\cite{sun2024shmt}                  & 0.932 & 0.714 & 5.87 & 4.22 & 0.937 & 0.716 & 4.78 & 5.00 \\
    StableMakeup~\cite{zhang2025stablemakeup}          & \underline{0.959} & \textbf{0.809} & 8.34 & \underline{7.30} & \textbf{0.967} & \textbf{0.836} & 7.81 & 7.89 \\
    Banana~Pro~\cite{deepmind_gemini_image_pro}            & 0.952 & 0.726 & 8.34 & 7.07 & 0.955 & 0.756 & 8.18 & 7.71 \\
    GPT~1.5~\cite{openai_gpt_image_15_model}         & \textbf{0.963} & 0.739 & \underline{8.69} & \textbf{7.51} & \underline{0.964} & 0.748 & \underline{8.85} & \textbf{8.13} \\
    \textbf{Ours}         & 0.946 & 0.736 & \textbf{8.96} & 7.12 & 0.955 & \underline{0.778} & \textbf{9.14} & \underline{8.04} \\
    \midrule
    & \multicolumn{4}{c|}{\textbf{MT-Wild Test Set~\cite{jiang2020psgan}}} & \multicolumn{4}{c}{\textbf{Our MF2K Test Set}} \\
    \cmidrule(lr){2-5}\cmidrule(lr){6-9}
    \multirow{-2.6}{*}{\makecell[c]{\textbf{Methods}}}  &
    CLIP-I$\uparrow$ &
    DINO-I$\uparrow$ &
    MSim$_G$$\uparrow$ &
    MSim$_Q$$\uparrow$ &
    CLIP-I$\uparrow$ &
    DINO-I$\uparrow$ &
    MSim$_G$$\uparrow$ &
    MSim$_Q$$\uparrow$ \\
    \midrule
    PSGAN~\cite{jiang2020psgan}                 & 0.940 & 0.656 & 3.65 & 2.71 & 0.931 & 0.606 & 1.49 & 0.74 \\
    EleGANt~\cite{yang2022elegant}               & 0.942 & 0.687 & 5.30 & 4.30 & 0.937 & 0.622 & 2.60 & 1.87 \\
    MAD~\cite{ruan2025mad}                   & 0.941 & 0.661 & 3.93 & 2.85 & 0.930 & 0.643 & 1.86 & 1.37 \\
    SHMT~\cite{sun2024shmt}                  & 0.935 & 0.643 & 3.11 & 2.45 & 0.933 & 0.628 & 3.47 & 2.50 \\
    StableMakeup~\cite{zhang2025stablemakeup}          & \textbf{0.969} & \textbf{0.820} & 8.21 & \textbf{7.62} & \textbf{0.964} & \textbf{0.787} & 6.73 & 7.04 \\
    Banana~Pro~\cite{deepmind_gemini_image_pro}            & 0.965 & 0.768 & 8.45 & 7.39 & 0.957 & \underline{0.737} & 8.34 & 7.06 \\
    GPT~1.5~\cite{openai_gpt_image_15_model}         & \underline{0.968} & \underline{0.777} & \underline{8.74} & \underline{7.53} & \underline{0.964} & 0.653 & \underline{8.43} & \textbf{7.69} \\
    \textbf{Ours}         & 0.956 & 0.734 & \textbf{8.77} & 7.00 & 0.957 & 0.737 & \textbf{9.22} & \underline{7.68} \\
    \bottomrule
    \end{tabular}
\end{table*}

\begin{table*}[!t]
    \centering
    \caption{Makeup similarity (MSim) under two additional, more recent VLM judges (Qwen3.6-Plus and Kimi K2.6) on the four strongest methods.
    Best and second best are in \textbf{bold} and \underline{underline}.}
    \label{tab:supp_more_vlm}
    \scriptsize
    \setlength{\tabcolsep}{5pt}
    \renewcommand{\arraystretch}{1.1}
    \begin{tabular}{l|cccc|cccc}
    \toprule
    & \multicolumn{4}{c|}{\textbf{Qwen3.6-Plus}} & \multicolumn{4}{c}{\textbf{Kimi K2.6}} \\
    \cmidrule(lr){2-5}\cmidrule(lr){6-9}
    \multirow{-2.6}{*}{\textbf{Methods}} & MT & LADN & MT-Wild & MF2K & MT & LADN & MT-Wild & MF2K \\
    \midrule
    StableMakeup~\cite{zhang2025stablemakeup} & \underline{7.51} & 7.88 & \textbf{8.19} & 6.95 & 6.79 & 6.64 & 7.20 & 5.88 \\
    Banana~Pro~\cite{deepmind_gemini_image_pro} & 6.94 & 7.61 & 7.42 & 7.74 & 6.55 & 6.79 & 6.97 & 6.74 \\
    GPT~1.5~\cite{openai_gpt_image_15_model} & 7.28 & \underline{8.03} & \underline{8.03} & \underline{7.95} & \underline{6.93} & \textbf{7.32} & \textbf{7.54} & \underline{7.24} \\
    \textbf{Ours} & \textbf{7.58} & \textbf{8.22} & 7.27 & \textbf{8.32} & \textbf{7.09} & \underline{7.13} & \underline{7.51} & \textbf{7.32} \\
    \bottomrule
    \end{tabular}
\end{table*}

Here we report these conventional feature-based metrics in Tab.~\ref{tab:supp_feature_metrics}.
To address the evaluation gap for localized fine-grained details, we further employ rubric-based VLMs as more interpretable and human-aligned judges.
Our two primary metrics are MSim$_G$, scored by a leading closed-source model (Gemini 3 Pro~\cite{deepmind_gemini_3_pro}), and MSim$_Q$, scored by a open-source model (Qwen2.5-VL-72B-Instruct~\cite{bai2025qwen25vltechnicalreport}).
To further corroborate these scores with the latest models, we additionally report two more recent VLMs, Qwen3.6-Plus and Kimi K2.6, as supplementary judges (Tab.~\ref{tab:supp_more_vlm}).
The reasoning logs in Fig.~\ref{fig:supp_metric_failures} show that these VLMs explicitly assess key cosmetic characteristics, local details, and style drifts, aligning closely with human perception.
Across these judges, ART ranks first on 9 and within the top two on 13 of the 16 method--judge settings, indicating that its advantage is consistent across judges and reflects genuine cosmetic fidelity.

\FloatBarrier

\noindent\begin{minipage}{\linewidth}
\noindent \textbf{VLM Evaluation Prompt.} To ensure reproducibility, the instruction prompt utilized for our VLM-as-a-Judge protocol is provided below:

\begin{lstlisting}
You are an evaluator for a makeup transfer task. You are given two images:
1) Reference makeup image (the first image)
2) Makeup output image (the second image)
Your job is to score how similar the makeup in the output is to the reference.
Ignore whether the person is the same, and ignore hair/background differences unless they prevent you from judging the makeup.
Return ONE score only: makeup_similarity_score in [0, 10], using 0.5 increments.

Score using these three questions (in order of importance):
1) Do the KEY makeup characteristics match? "Key characteristics" means the most visually dominant makeup elements in the reference, which can be ANY of:
   - eye makeup, lip makeup, base/skin tone and finish, blush/contour/highlight,
   - graphic/face-paint/SFX elements (lines, cracks, geometric shapes, special textures), etc.
2) Do the LOCAL details mostly align? Check placement, shape, intensity, color tone, edge cleanliness, and whether any important parts are missing.
3) Is there any obvious hallucination or style drift? Penalize if the output adds a strong makeup element not present in the reference, or changes the overall style into a different look.

Use the following score bands (do NOT create sub-scores):
- 9.0-10.0: Key characteristics match very closely; details mostly correct; little/no hallucination or drift.
- 7.0-8.5: Key characteristics match; some noticeable detail differences or weakening; still clearly the same look.
- 5.0-6.5: Partially similar; key elements missing or clearly off; or moderate drift/hallucination.
- 3.0-4.5: Low similarity; many key elements wrong/missing; or strong hallucination/drift.
- 0.0-2.5: Not matching; key elements absent or completely different style.

Output JSON only:
{
  "reasoning": "1-3 short sentences: what the key characteristics are in the reference, whether they match in the output, and the main mismatches/hallucinations.",
  "makeup_similarity_score": <number>
}
\end{lstlisting}
\end{minipage}
\FloatBarrier

\noindent \textbf{Evaluation on a Larger Test Set.} To confirm that our findings are not specific to the 100-pair test set, we further evaluate on a larger LADN set of 1{,}000 source--reference pairs.
As reported in Tab.~\ref{tab:supp_ladn_1k}, ART's standing is unchanged from the 100-pair protocol: it still ranks first in MSim$_G$, ID, and L2-M and second in MSim$_Q$, confirming that our evaluation is stable at a larger scale.

\begin{table}[!t]
    \centering
    \setlength{\belowcaptionskip}{4pt}
    \caption{Evaluation on a larger LADN test set~\cite{Gu_2019_ICCV} with 1{,}000 source--reference pairs, on the four strongest methods.
    ART's rankings stay consistent with the standard 100-pair protocol.
    Best and second best are in \textbf{bold} and \underline{underline}.}
    \label{tab:supp_ladn_1k}
    \scriptsize
    \setlength{\tabcolsep}{8pt}
    \renewcommand{\arraystretch}{1.1}
    \begin{tabular}{l|ccccc}
    \toprule
    \textbf{Methods} & MSim$_G$$\uparrow$ & MSim$_Q$$\uparrow$ & ID$\uparrow$ & L2-M$\downarrow$ & FID$\downarrow$ \\
    \midrule
    StableMakeup~\cite{zhang2025stablemakeup} & 7.73 & 7.74 & 0.45 & \underline{9.41} & \textbf{63.27} \\
    Banana~Pro~\cite{deepmind_gemini_image_pro} & 8.23 & 7.55 & \underline{0.63} & 39.82 & 71.84 \\
    GPT~1.5~\cite{openai_gpt_image_15_model} & \underline{8.76} & \textbf{8.24} & 0.33 & 46.03 & \underline{67.41} \\
    \textbf{Ours} & \textbf{9.08} & \underline{8.18} & \textbf{0.80} & \textbf{3.82} & 69.55 \\
    \bottomrule
    \end{tabular}
    \vspace{-8pt}
\end{table}

\subsection{User Study Details}
\label{subsec:user_study_details}
\noindent \textbf{Evaluation Protocol and Demographics.} To comprehensively assess human perceptual preferences, we conducted a randomized blind test, collecting $864$ valid ratings from $21$ diverse participants (66.7\% male, 33.3\% female; ages 18 to 45+).
Ensuring strict evaluation standards, the cohort possessed highly relevant expertise: $71.4\%$ were AI/CV researchers, $42.9\%$ had photography backgrounds, and $28.6\%$ were familiar with cosmetics.
Outputs from our method and the baselines were randomly shuffled to prevent subjective bias.

\noindent \textbf{Evaluation Rubric.} Participants rated each randomized output on a 1--5 Likert scale (1 = Very Poor, 5 = Excellent) across three dimensions:
\begin{itemize}
    \item \textbf{Makeup Similarity:} Assesses cosmetic fidelity, ranging from mismatched (1) to near-identical preservation of key elements and details (5).
    \item \textbf{Identity Consistency:} Evaluates structural preservation of the source, from unrecognizable distortion (1) to stable, identical facial features (5).
    \item \textbf{Image Quality:} Measures overall realism, ranging from severe hallucinatory artifacts (1) to clear, natural, and highly credible visuals (5).
\end{itemize}

\begin{table*}[!t]
    \centering
    \caption{Detailed user study breakdown across four individual test sets.
    Our method consistently ranks first or second in virtually all aspects, demonstrating unparalleled robustness and identity preservation compared to commercial generation models.}
    \label{tab:supp_user_study_breakdown}
    \scriptsize
    \setlength{\tabcolsep}{2.5pt}
    \renewcommand{\arraystretch}{1.0}
    \begin{tabular}{l|ccc|ccc}
    \toprule
    & \multicolumn{3}{c|}{\textbf{MT Test Set~\cite{li2018beautygan}}} & \multicolumn{3}{c}{\textbf{LADN Test Set~\cite{Gu_2019_ICCV}}} \\
    \cmidrule(lr){2-4}\cmidrule(lr){5-7}
    \multirow{-2.6}{*}{\makecell[c]{\textbf{Methods}}}  &
    \makecell[c]{Makeup \\ Similarity $\uparrow$} &
    \makecell[c]{ID \\ Consistency $\uparrow$} &
    \makecell[c]{Image \\ Quality $\uparrow$} &
    \makecell[c]{Makeup \\ Similarity $\uparrow$} &
    \makecell[c]{ID \\ Consistency $\uparrow$} &
    \makecell[c]{Image \\ Quality $\uparrow$} \\
    \midrule
    PSGAN~\cite{jiang2020psgan}            & 1.82 & 3.18 & 2.27 & 2.18 & 3.27 & 2.36 \\
    EleGANt~\cite{yang2022elegant}          & 2.41 & 3.27 & 2.68 & 2.77 & \underline{3.59} & 2.77 \\
    MAD~\cite{ruan2025mad}             & 2.55 & 3.14 & 2.86 & 1.27 & 1.77 & 1.23 \\
    SHMT~\cite{sun2024shmt}             & 2.18 & \underline{3.32} & 2.32 & 2.00 & 3.55 & 2.41 \\
    StableMakeup~\cite{zhang2025stablemakeup}     & 2.64 & 2.36 & 2.91 & 3.32 & 2.41 & 2.86 \\
    Banana~Pro~\cite{deepmind_gemini_image_pro}  & \underline{3.14} & 2.95 & \underline{3.64} & \textbf{3.68} & 3.32 & \underline{4.00} \\
    GPT~1.5~\cite{openai_gpt_image_15_model}    & 2.95 & 2.14 & \textbf{3.73} & 3.32 & 2.41 & 3.77 \\
    \textbf{Ours}    & \textbf{3.32} & \textbf{3.95} & 3.55 & \underline{3.64} & \textbf{4.05} & \textbf{4.05} \\
    \midrule
    & \multicolumn{3}{c|}{\textbf{MT-Wild Test Set~\cite{jiang2020psgan}}} & \multicolumn{3}{c}{\textbf{MF2K Test Set}} \\
    \cmidrule(lr){2-4}\cmidrule(lr){5-7}
    \multirow{-2.6}{*}{\makecell[c]{\textbf{Methods}}}  &
    \makecell[c]{Makeup \\ Similarity $\uparrow$} &
    \makecell[c]{ID \\ Consistency $\uparrow$} &
    \makecell[c]{Image \\ Quality $\uparrow$} &
    \makecell[c]{Makeup \\ Similarity $\uparrow$} &
    \makecell[c]{ID \\ Consistency $\uparrow$} &
    \makecell[c]{Image \\ Quality $\uparrow$} \\
    \midrule
    PSGAN~\cite{jiang2020psgan}            & 1.36 & 3.09 & 2.18 & 1.33 & 2.71 & 1.76 \\
    EleGANt~\cite{yang2022elegant}          & 2.00 & 2.77 & 2.18 & 1.40 & 2.74 & 1.74 \\
    MAD~\cite{ruan2025mad}             & 1.45 & 1.91 & 1.23 & 1.12 & 1.50 & 1.10 \\
    SHMT~\cite{sun2024shmt}             & 1.32 & \underline{3.55} & 2.23 & 1.57 & 2.81 & 1.98 \\
    StableMakeup~\cite{zhang2025stablemakeup}     & 2.82 & 2.55 & 2.77 & 2.48 & 2.02 & 2.55 \\
    Banana~Pro~\cite{deepmind_gemini_image_pro}  & \textbf{3.50} & 2.82 & \textbf{4.00} & 3.40 & \underline{3.26} & \underline{3.69} \\
    GPT~1.5~\cite{openai_gpt_image_15_model}    & 3.23 & 1.82 & \underline{3.68} & \underline{3.55} & 2.19 & \textbf{3.98} \\
    \textbf{Ours}    & \underline{3.36} & \textbf{3.95} & 3.64 & \textbf{4.02} & \textbf{4.10} & \textbf{3.98} \\
    \bottomrule
    \end{tabular}
\end{table*}

\noindent \textbf{Quantitative Results.} While the overall user study summary is discussed in the main paper, Tab.~\ref{tab:supp_user_study_breakdown} provides a detailed breakdown of the perceptual scores across four individual evaluation sets.
Consistent with the overall trend, our method demonstrates consistent robustness across all subsets.
Notably, while closed-source commercial models (Nano Banana Pro~\cite{deepmind_gemini_image_pro} and GPT Image 1.5~\cite{openai_gpt_image_15_model}) achieve competitive image quality, they consistently suffer from identity drift.
In contrast, our method excels in fine-grained cosmetic reconstruction while maintaining the highest identity preservation across all evaluated scenarios.

\section{Limitations and Failure Cases}
\label{sec:suppl_limitations}

\begin{figure}[!t]
    \centering
    \includegraphics[width=\linewidth]{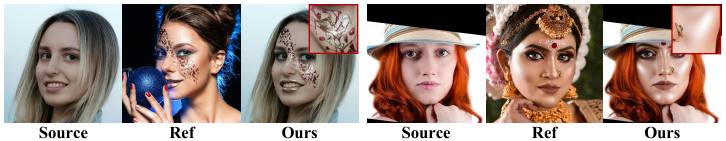}
    \caption{{Failure case.} The model accurately captures the metallic texture and specular highlights from the reference image.
    These transferred reflections appear physically unnatural under the soft and diffuse lighting environment of the source portrait.}
    \label{fig:supp_failure}
\end{figure}

Fig.~\ref{fig:supp_failure} illustrates a failure case involving a significant illumination mismatch between the source and reference images.
The reference portrait is captured under strong directional lighting, producing a distinct glittery appearance on the cosmetics.
Our generated output faithfully retains these strong specular highlights and metallic textures.
However, such intense reflections physically contradict the soft and diffuse lighting environment of the source image.
This reveals that our current framework transfers visual appearance cues without explicitly enforcing physical lighting and reflectance consistency.